\documentclass[10pt,journal,cspaper,compsoc]{IEEEtran}
\usepackage{epsfig}
\usepackage{titlesec}
\usepackage{float}
\usepackage{graphicx}
\usepackage{tabularx}
\usepackage{threeparttable}
\usepackage{amsmath}
\usepackage{amssymb}
\usepackage{amsthm}
\usepackage{caption}
\usepackage[ruled]{algorithm2e}
\usepackage{threeparttable}
\usepackage{colortbl,booktabs}
\usepackage{threeparttable}
\usepackage{subeqnarray}
\usepackage{cases}
\usepackage{multirow}
\usepackage{pifont}
\usepackage{array}
\usepackage{bm}
\usepackage{ragged2e}
\usepackage[switch]{lineno}
\usepackage{threeparttable}
\usepackage{fixltx2e}
\usepackage{enumerate}
\usepackage{epstopdf}
\usepackage{multirow}
\usepackage{graphicx}
\usepackage{float}
\usepackage{subfigure}
\usepackage{algorithmic}
\usepackage{amsmath}
\usepackage[hyphenbreaks]{breakurl}
\usepackage[hyphens]{url}
\usepackage{cite}

\newcolumntype{I}{!{\vrule width 2pt}}
\newlength\savedwidth

\newlength\savewidth

\theoremstyle{definition}

\begin{document}
\title{MGAA: Multi-Granular Adaptive Allocation  for  Low-Rank Compression of LLMs}

\bibliographystyle{IEEEtran}
    
\author{Guangyan Li,
        Yongqiang Tang
        and Wensheng Zhang
\thanks {G. Li is with the State Key Laboratory of Multimodal Artificial Intelligence Systems, Institute of Automation, Chinese Academy of Sciences, Beijing 100190, China, and also with the School of Artificial Intelligence, University of Chinese Academy of Sciences, Beijing 100049, China. (email: liguangyan2022@ia.ac.cn)}
\thanks { Y. Tang is with the State Key Laboratory of Multimodal Artificial Intelligence Systems, Institute of Automation, Chinese Academy of Sciences, Beijing, 100190, China. (e-mail: yongqiang.tang@ia.ac.cn)}
\thanks { W. Zhang is with the State Key Laboratory of Multimodal Artificial Intelligence Systems, Institute of Automation, Chinese Academy of Sciences, Beijing, 100190, China, and also with the School of Computer Science and Cyber Engineering, Guangzhou University, Guangzhou, 510006, China. (email: zhangwenshengia@hotmail.com)}
}


\IEEEtitleabstractindextext{
\begin{abstract}
\justifying
The enormous parameter scale of large language models (LLMs) has made model compression a research  hotspot, which aims to alleviate computational resource demands during deployment and inference. As a promising direction, low-rank approximation technique has made remarkable achievements. Nevertheless, unfortunately, the vast majority of studies to low-rank approximation compression generally apply uniform compression ratios across all weight matrices, while disregarding their inherently differentiated impacts on the model's performance. Although a few recent work attempts to employ heuristic search strategies to achieve the optimal parameter allocation,  such strategies are computationally inefficient and lose the generalization ability in the era of LLMs. 
In this study, we propose a novel parameter Multi-Granular Adaptive Allocation (MGAA) method, which can adaptively allocate parameters between and within sublayers without task-specific evaluations in the compression process. MGAA consists of two components: 1) Among different sublayers,  it assigns compression ratios based on their cosine similarity between inputs and outputs,  allowing for a more tailored compression in sublayers with varying degrees of importance, and 2) Within each sublayer, it allocates different compression ratios to weight matrices based on their energy distribution characteristics, ensuring a consistent energy retention ratio while optimizing compression efficiency. 
Comprehensive evaluations of MGAA across multiple LLMs backbone models and benchmark datasets demonstrate its superior performance. Additionally, we apply our MGAA to multimodal model LLaVA,  exhibiting remarkable performance improvements. 
More importantly, MGAA could serve as a versatile plug-and-play solution that can be seamlessly integrated with various low-rank approximation techniques. Its consistent performance improvements across  existing  low-rank compression methods further validate its broad applicability and strong generalization capability.
\end{abstract}
\begin{IEEEkeywords}
low-rank approximation, model compression, transformer model, adaptive allocation, plug-and-play method
\end{IEEEkeywords}
}

\maketitle

\IEEEdisplaynontitleabstractindextext

%


\section{Introduction}

\IEEEPARstart{T}{he} advent of Large Language Models (LLMs) \cite{LLaMA,OPT,Mistral} has revolutionized natural language processing, achieving unprecedented performance across various tasks \cite{chain}. However, this remarkable performance comes with substantial computational costs: modern LLMs require hundreds of billions of parameters and massive computational resources for inference. Such resource demands pose significant challenges for real-world deployment, particularly in resource-constrained environments and edge devices.

To address these computational challenges, model compression for LLMs has emerged as a critical research direction. Current approaches primarily fall into four categories: 1) quantization \cite{gptq,AWQ,smoothquant,int8,pami_quant1}, which reduces numerical precision while maintaining model performance; 2) pruning \cite{llmpruner,flap,slicegpt,pami_prune1,pami_prun2}, which eliminates redundant parameters; 3) knowledge distillation \cite{teacher,SCOTTSC,pami_distill1}, which transfers knowledge to smaller models; and 4) low-rank approximation \cite{AFM,ASVD,SVD-llm,pami_low_rank1}, which decomposes weight matrices into lower-rank matrices. These approaches offer unique advantages and challenges in terms of compression ratio, inference speed, and deployment complexity. Among these approaches, low-rank approximation has gained significant attention due to its unique combination of theoretical guarantees and task-agnostic nature. This method not only provides clear mathematical insights into the compression process but also achieves substantial parameter reduction.

The application of low-rank approximation to LLMs has evolved through several key stages. Initial efforts focused on basic matrix decomposition, where SVD was applied to compress BERT models \cite{first_svd}. Subsequently, task-specific approaches emerged: Drone \cite{DRONE} introduced simultaneous decomposition of weight matrices and input activations, while FWSVD \cite{FWSVD} leveraged task-specific gradient information to identify and preserve critical weights. However, these methods' dependence on specific tasks limited their applicability to LLMs, where preserving general capabilities is crucial. To address this limitation, recent methods have shifted towards task-agnostic approaches. ASVD \cite{ASVD}, AWSVD \cite{lorap}, and SVD-LLM \cite{SVD-llm} utilize task-agnostic input activations to weight matrices before applying low-rank approximation. In contrast to these weight-space methods, AFM \cite{AFM} proposed a novel feature-space compression paradigm using PCA on output features, demonstrating scalability through successful compression of a 16B model in LORD \cite{lord}.

Despite the significant advances in low-rank approximation compression techniques, \emph{the challenge of efficient parameter allocation across the entire model} during compression remains largely unaddressed. Actually, recent studies \cite{llmpruner,shortgpt,ASVD,AFM} have revealed that different layers and weight matrices within the model exhibit varying degrees of importance to model performance. This observation suggests that a differentiated parameter allocation strategy could substantially enhance both the performance and generalization capability of compressed model. 
Unfortunately, current research to low-rank approximation compression typically apply uniform compression ratios across all weight matrices, while disregarding their inherent differences in importance. More recently, a few works \cite{AFM,ASVD} attempt to employ the heuristic search strategy, which determine differentiated compression ratios  based on performance degradation in downstream validation tasks: weight matrices inducing larger performance drops receive lower compression ratios, while those causing minimal impact are compressed more aggressively. Despite the impressive performance improvement, this approach presents two critical limitations. First, it requires extensive evaluations on validation sets, making it computationally prohibitive for LLMs. Second, the resulting compression strategy becomes inherently biased towards specific validation tasks, potentially compromising the model's general capabilities, particularly its zero-shot performance across diverse tasks. 

To address these limitations, we propose a novel parameter Multi-Granularity Adaptive Allocation (MGAA)  method for low-rank compression,  which operates at both inter- and intra-sublayer levels. At the sublayer level, 
we employ cosine similarity to estimate the importance of each sublayer and strategically allocate compression ratios accordingly. This compression strategy preserves the model's critical components during compression, allowing the compressed model to better maintain its performance. At the intra-sublayer level, we find that 
model performance remains consistent when different matrices maintain equivalent energy retention ratios. This insight leads us to propose an energy-balanced parameter allocation strategy based on energy distribution. We assign different compression ratios to weight matrices while ensuring consistent energy retention, determined by the eigenvalue distribution of each weight matrix. 
Note that, compared with the present parameter allocation rivals \cite{AFM,ASVD},
our MGAA owns several distinct advantages. First, MGAA effectively reduces compression time by only leveraging  activation values with just two forward propagation. 
Besides, by using task-agnostic calibration data, MGAA avoids overfitting to specific downstream tasks, ensuring robust performance across diverse tasks and improving accuracy in zero-shot learning scenarios. More importantly, MGAA can serve as a plug-and-play parameter allocation module that can be directly combined with various low-rank approximation techniques. 

Our main contributions are summarized as follows:
\begin{itemize}
    \item  We propose a sublayer-level compression ratio allocation strategy based on the cosine similarity between input and output features. This strategy adaptively allocates compression ratios to sublayers according to their importance, preserving the critical components of the model during compression.

    \item Within each sublayer, we introduce an energy-balanced parameter allocation method based on energy distribution. This approach, founded on the relationship between energy preservation ratio and performance, ensures consistent energy retention proportions across different matrices during compression, resulting in  superior  performance.

     \item  We conduct comprehensive evaluations across multiple models (including LLaMA1, Vicuna, LLaMA3, Mistral) using zero-shot perplexity on the WikiText2, PTB, and C4 datasets, as well as seven common sense reasoning datasets. Additionally, we extend our approach to multimodal model LLaVA, which exhibits remarkable performance improvements across various compression ratios.

    \item MGAA significantly reduces compression time by leveraging only activation values from the forward pass, achieving compression in just two forward propagations on calibration datasets. It avoids overfitting to specific downstream tasks by using task-agnostic calibration data, ensuring robust performance across diverse tasks and improving zero-shot learning accuracy. Additionally, MGAA serves as a plug-and-play parameter allocation method, compatible with various low-rank approximation techniques, and enhances performance when applied to existing methods like AFM\cite{AFM} and ASVD\cite{ASVD}.
\end{itemize}

\section{Background}
\subsection{Transformer Model}
The Transformer architecture consists of multiple layers, each comprising two primary sublayers: Multi-Head Attention (MHA) and Feed-Forward Network (FFN). For a given input $\mathbf{X} \in \mathbb{R}^{d_{\text{in}} \times L}$ (where $d_{\text{in}}$ is the input dimension and $L$ is the sequence length) to a sublayer and its corresponding output $\mathbf{Y} \in \mathbb{R}^{d_{\text{out}} \times L}$, the computation can be formally expressed as:
\begin{align}\label{sublayer expression}
\mathbf{Y}= \mathrm{LAYER}(\mathrm{LN}(\mathbf{X}))+\mathbf{X},
\end{align}
where LAYER denotes either MHA or FFN, and LN represents layer normalization. Each sublayer contains distinct weight matrices serving different functionalities. Specifically, the MHA sublayer encompasses four weight matrices:  $\mathbf{W}_{q}$, $\mathbf{W}_{k}$, $\mathbf{W}_{v}$, and $\mathbf{W}_{o}$ for query, key, value, and output projections, respectively. The FFN sublayer consists of $\mathbf{W}_{U}$, $\mathbf{W}_{D}$, and potentially $\mathbf{W}_{G}$ for gating mechanisms. Throughout this paper, we define matrix multiplication as $\mathbf{y} = \mathbf{W} \mathbf{x}$, where $\mathbf{W} \in \mathbb{R}^{d_{\text{out}}\times d_{\text{in}}}$ represents any weight matrix in the model.

\subsection{Low-Rank Approximation}
Given the immense scale of LLMs, compressing the entire model simultaneously poses significant challenges. Recent studies \cite{gptq,OBC,sparsegpt} address this challenge by decomposing the global compression task into multiple independent matrix compression problems. This matrix-wise compression strategy not only reduces memory requirements but also enables efficient parallel computation.
For each weight matrix $\mathbf{W} \in \mathbb{R}^{d_{\text{out}} \times d_{\text{in}}}$, the most straightforward compression approach is direct weight approximation:
\begin{equation}\label{weight svd}
  {\underset{\mathbf{L},\mathbf{R}}{\mathrm{min}}} \ \ \|\mathbf{W}-\mathbf{LR}\|^{2}_{F}.
\end{equation}
where $\mathbf{L} \in \mathbb{R}^{d_{\text{out}} \times r}$ and $\mathbf{R} \in \mathbb{R}^{r \times d_{\text{in}}}$ are low-rank matrices. The optimal solution is obtained through  $\text{SVD}(\mathbf{W})=\mathbf{U}\mathbf{\Sigma}\mathbf{V}$, where $\mathbf{L}=\mathbf{U}\mathbf{\Sigma}$ and $\mathbf{R}=\mathbf{V}$.
Recent methods have explored more advanced approaches by considering additional information during compression, leading to two primary approaches: weighted decomposition and feature decomposition. Given input data $\mathbf{X} \in \mathbb{R}^{d_{\text{in}} \times L}$ and an invertible importance matrix $\mathbf{S} \in \mathbb{R}^{d_{\text{in}} \times d_{\text{in}}}$, weighted decomposition aims to solve:
\begin{equation}\label{layer-wise-question}
  {\underset{\mathbf{L},\mathbf{R}}{\mathrm{min}}} \ \ \|\mathbf{WS}-\mathbf{LRS}\|^{2}_{F},
\end{equation}
where $||\cdot||_{F}$ denotes the Frobenius norm. The optimal solution is obtained through $\text{SVD}(\mathbf{WS})=\mathbf{U} \mathbf{\Sigma} \mathbf{V}$, yielding $\mathbf{L}=\mathbf{U}\mathbf{\Sigma}$ and $\mathbf{R} = \mathbf{VS}^{-1}$.
Feature decomposition methods, such as AFM \cite{AFM}, focus on the output feature space:
\begin{equation}\label{layer-wise}
  {\underset{\mathbf{L},\mathbf{R}}{\mathrm{min}}} \ \ \|\mathbf{WX}-\mathbf{LRX}\|^{2}_{F},
\end{equation}
where the solution yields low-rank matrices that optimally preserve the model's feature information. For all these methods, the rank $r$ is determined by compression ratio $p$ as:
\begin{equation}\label{cal rank}
r = \frac{d_{\text{in}} \cdot d_{\text{out}} \cdot (1-p)}{d_{\text{in}} + d_{\text{out}}}.
\end{equation}

\subsection{Heuristic Parameter Allocation} 
The most common heuristic approach for parameter allocation relies on sensitivity of different matrices. These approachs measure the sensitivity of each matrix  to compress by evaluating the model's behavior changes after compression. Specifically, given a validation set $\mathcal{D}$, the sensitivity score for matrix $\mathbf{W}_{i}$ under retained rank $k$ can be measured as:
\begin{equation}
P_i(k) = \sum_{x\in\mathcal{D}} F(f(x,\mathbf{W}_{i}), f(x,\mathbf{W}_{i}(k)),
\end{equation}
where $f(x,\mathbf{W}_{i})$ denotes the model output with original parameters $\mathbf{W}_{i}$, and $f(x,\mathbf{W}_i(k))$ represents the model output where matrix $\mathbf{W}_i$ is compressed to rank $k$ while keeping other matrices unchanged. The function $F(\cdot,\cdot)$ measures the output discrepancy between original and compressed models, which can be instantiated as either task-specific performance difference $F(y_1,y_2) = y_1 - y_2$ (e.g., accuracy for classification, BLEU score for translation) or distribution divergence $F(y_1,y_2) = D(y_1\|y_2)$ (e.g., KL-divergence, Wasserstein distance, Jensen-Shannon divergence). The resulting sensitivity score $P_i(k)$ quantifies how sensitive matrix $\mathbf{W}_i$ is when compressed to rank $k$, with higher values indicating the matrix is more crucial for model functionality. 
Based on these sensitivity scores, the parameter allocation problem is formulated as an optimization problem:
\begin{equation}
\begin{aligned}
\min_{\{k_i\}_{i=1}^N} & \sum_{i=1}^N P_i(k_i), \ \ 
\text{s.t.} & \sum_{i=1}^N k_i \leq R_{\text{tar}},
\end{aligned}
\end{equation}
where $N$ denotes the total number of weight matrices to be compressed, $k_i$ represents the retained rank for matrix $\mathbf{W}_i$, and $R_{\text{tar}}$ is the target total rank budget. This optimization aims to minimize the sum of sensitivity scores while ensuring the total retained ranks do not exceed the budget constraint.

While this approach provides a systematic way to allocate parameters, it has several limitations. Despite sampling ranks at fixed intervals (e.g., multiples of 32) to reduce computation, it still requires extensive model evaluations to compute sensitivity scores. Moreover, these sensitivity scores inherently depend on specific validation tasks, potentially compromising the model's general capabilities. Additionally, the discrete nature of ranks and independence assumption between matrices typically lead to suboptimal solutions through greedy algorithms. These limitations motivate our development of an efficient and task-agnostic parameter allocation strategy based on inherent matrix properties rather than empirical sensitivity measures.

\section{Method}
\begin{figure*}[h]
  \centering
    \subfigure[Sublayer-Wise Compression Ratio Allocation]{\includegraphics[width=0.48\textwidth]{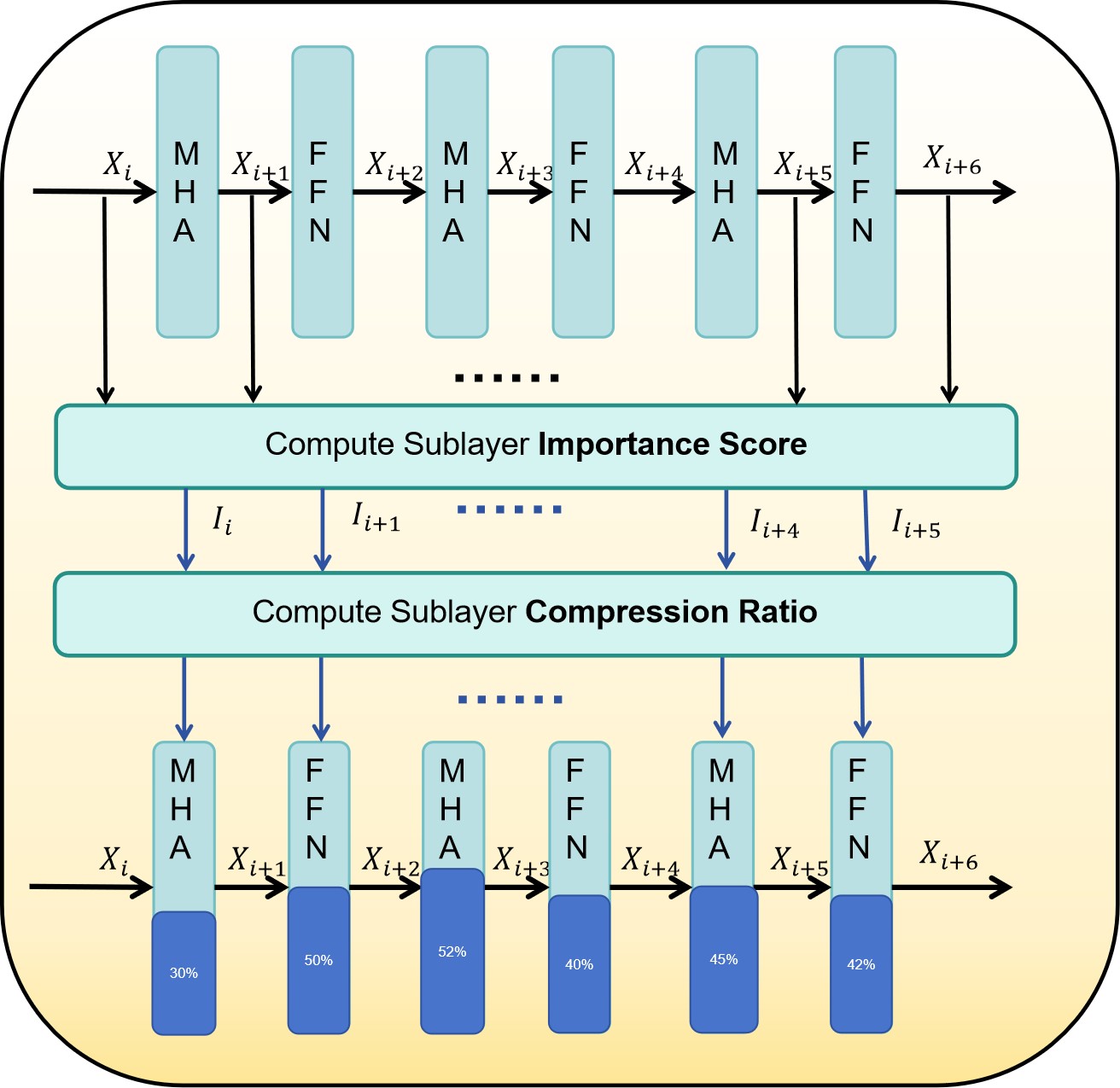}}
    \hspace{5pt} 
    \subfigure[Energy-Balanced Matrix Parameter Allocation]{\includegraphics[width=0.48\textwidth]{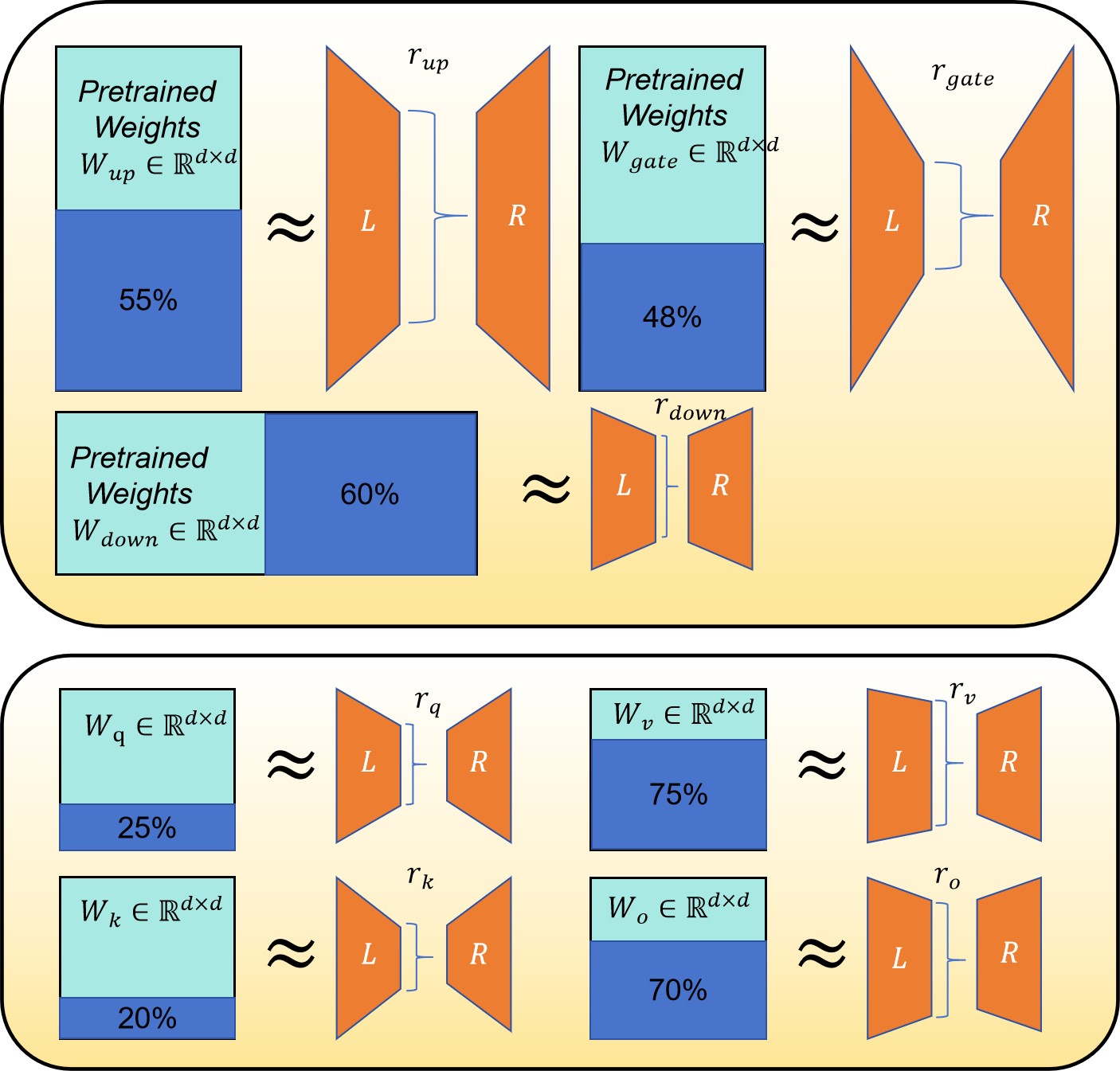}}
  \caption{An overview of multi-granular adaptive allocation.The left figure illustrates the allocation of compression ratios to sublayers based on the eigenvalues of their input and output features. The right figure depicts the preservation of different ranks for various weight matrices based on energy balancing.}
  \label{main}
\end{figure*}

In this section, we first analyze existing PCA compression methods and their approximation properties. Then, we explain the key components and mechanism of our framework for adaptive parameter allocation. Finally, we present the detailed allocation strategy and algorithm implementation, which is illustrated in Figure \ref{main}.

\subsection{Low-Rank Approximation via PCA}
Performing decomposition in the output feature space allows  more efficient utilization of calibration data. Hence, we opt for principal component analysis (PCA), i.e., Eq. (\ref{layer-wise}), as the foundational method to illustrate our MGAA. Given the weight matrix \( \mathbf{W} \in \mathbb{R}^{ d_{\text{out}} \times d_{\text{in}}} \), input data $ \mathbf{X} \in \mathbb{R}^{ d_{\text{in}} \times L} $, and the output data  $ \mathbf{Y} \in \mathbb{R}^{ d_{\text{out}} \times L} $, the PCA low-rank approximation method begins with the eigenvalue decomposition (EVD) of the autocorrelation matrix of the outputs. Specifically, the process is as follows:
\begin{equation}
\begin{aligned}\label{autocorrelation matrix}
  \mathbf{YY}^\mathrm{T}&=\mathbf{WX(WX)}^\mathrm{T} = \mathbf{WX}\mathbf{X}^\mathrm{T}\mathbf{W}^\mathrm{T} =\mathbf{U\Lambda U}^\mathrm{T}, 
\end{aligned}
\end{equation}
where $\mathbf{\Lambda}$ is a diagonal matrix containing $d_{out}$ eigenvalues arranged in descending order. It can be represented as:
\begin{equation}\label{x_din}
\mathbf{\Lambda}=\mathrm{diag}(\lambda_{1},\lambda_{2},\ldots,\lambda_{d_{out}}), \quad \lambda_{1} \geq \lambda_{2}\geq \ldots \geq 
 \lambda_{d_{out}}.
\end{equation}

Since \( \mathbf{YY}^\mathrm{T} \) is a symmetric matrix, eigenvector matrix \( \mathbf{U} \) is an orthogonal matrix and meets $\mathbf{UU}^\mathrm{T}=\mathbf{I}$.
After determining the retained rank \( r \) of current matrix by Eq. \eqref{cal rank}, we extract the first \( r \) eigenvectors from the matrix \( \mathbf{U} \) to form \( \mathbf{U_r} \in \mathbb{R}^{ d_{\text{out}} \times r }\). Since $\mathbf{U_r}$ consists of the orthonormal eigenvectors corresponding to the top-r largest eigenvalues, which preserve the most significant information in the data space, we have $\mathbf{U_r}\mathbf{U_r}^\top \approx \mathbf{I}$.
The truncated $k$ ($k=d_{out}-r$) eigenvectors form \( \mathbf{U_k}\in \mathbb{R}^{ d_{\text{out}} \times k }\). Subsequently, based on \( \mathbf{U_{r}} \), we decompose the weight matrix $\mathbf{W}$ into two low-rank matrices $\mathbf{L} \in \mathbb{R}^{ d_{\text{out}} \times r}$ and  $\mathbf{R} \in \mathbb{R}^{r \times d_{in}}$. The process is as follows:
\begin{equation}\label{compression}
\mathbf{Y}=\mathbf{WX} \approx \mathbf{U_{r}U_{r}}^\mathrm{T}\mathbf{WX} =\mathbf{LRX},
\end{equation}
where $\mathbf{L}=\mathbf{U_{r}},\mathbf{R}=\mathbf{U_{r}^\mathrm{T}\mathbf{W}}$. We represent the compression loss using the square of Frobenius norm of the output matrices before and after compression.
PCA low-rank approximation method can correlate the compression loss $\mathcal{L}$ with the truncated eigenvalues $\lambda_{i}(i>r)$. Using the Frobenius norm $\|\mathbf{A}\|_F = \left[\text{trace}(\mathbf{A}\mathbf{A}^\mathrm{T})\right]^{\frac{1}{2}}$ for a matrix $\mathbf{A}$, we can derive:

\begin{equation}\label{equal}
\begin{aligned}
\mathcal{L} &=\| \mathbf{UU}^\mathrm{T}\mathbf{WX}-\mathbf{U_{r}U_{r}}^\mathrm{T}\mathbf{WX} \|^{2}_{F} \\
 &=\mathrm{trace}(\mathbf{U_{k}\Lambda_{k}U_{k}^\mathrm{T}}) = \sum_{i=r+1}^{d_{out}} \lambda_i.
\end{aligned} 
\end{equation}

As indicated in Eq. (\ref{equal}), the compression loss in the PCA low-rank approximation method is equal to the sum of the truncated eigenvalues. By truncating the smallest eigenvalues and their corresponding eigenvectors, this method ensures minimal compression loss on the calibration dataset.

\subsection{Sublayer-Wise Compression Ratio Allocation}
In transformer architectures, different sublayers exhibit varying degrees of importance to the overall model performance. Estimating the importance of these sublayers efficiently and effectively remains a crucial challenge. According to Eq. (\ref{sublayer expression}), we can define the transformation of input features by sublayers as $T(\mathbf{X})$, which represents how the sublayer modifies the input features. When $T(\mathbf{X})$ is small, it indicates that the sublayer performs relatively simple feature transformations, preserving more characteristics of the original input features. This preservation suggests a higher degree of parameter redundancy within these layers, as simpler transformations typically require fewer parameters to accomplish their function. Conversely, when $T(\mathbf{X})$ demonstrates greater magnitude, it signifies more substantial feature transformations, indicating that the sublayer is learning more complex features and generating additional task-relevant information. This increased transformation complexity suggests these layers play a more significant role in the model's processing pipeline and are thus more critical for the model's overall functionality.

\begin{figure}[h]
  \centering
  \includegraphics[width=0.47\textwidth]{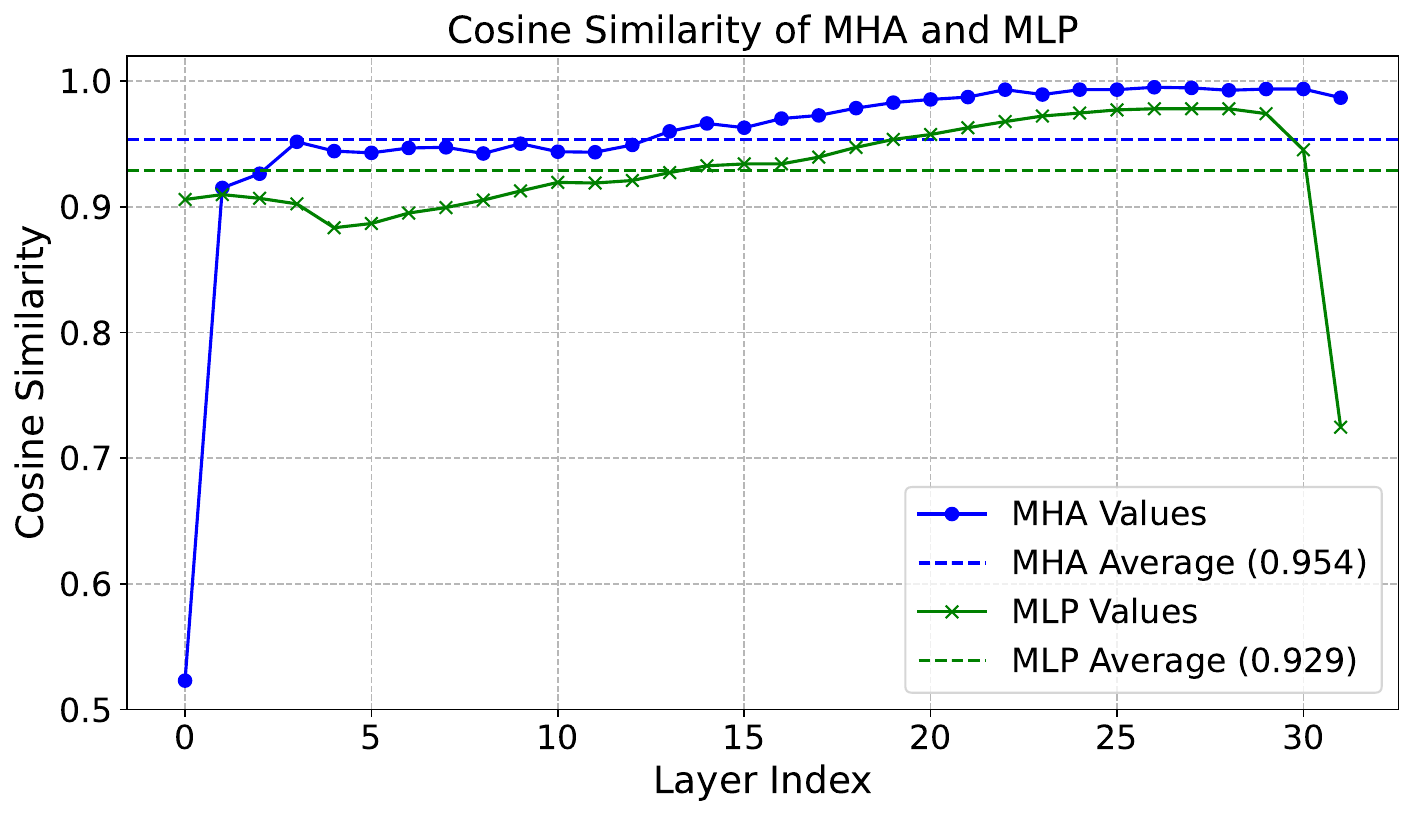}
  \vskip -0.07in
  \caption{The cosine similarity of different sublayers in the LLaMA1-7B model.}
  \label{Cosine similarity}
\end{figure}

Motivated by this theoretical analysis, we propose quantifying sublayer importance by examining the divergence between input and output features, as this divergence directly reflects the magnitude and complexity of the transformation $T(\mathbf{X})$. 
Specifically, we develop a cosine similarity-based importance measure that systematically quantifies each sublayer's contribution to the model's functionality. Given an input  $ \mathbf{X} \in \mathbb{R}^{d_{\text{in}} \times L} $ to a sublayer and its corresponding output  $ \mathbf{Y} \in \mathbb{R}^{d_{\text{out}} \times L} $,  we measure the importance of a sublayer by cosine similarity between the inputs and outputs of the sublayer, which is detailed as follows:
\begin{align}\label{importance}
   & I= \frac{1}{L}\sum_{m=1}^L\frac{\mathbf{X}_{m}^\mathrm{T} \mathbf{Y}_{m}}{\|\mathbf{X}_{m}\|_2 \|\mathbf{Y}_{m}\|_2}.
\end{align}
The subscript $m$ denotes the $m\text{-th}$ column of the respective matrix. 
The rationale behind this measure is that lower cosine similarity indicates more substantial feature transformation, suggesting higher importance of the sublayer in the network's computational process. Conversely, higher similarity implies less transformation of the input features, indicating lower importance. Based on this importance measure, we can determine appropriate compression ratios, allocating less compression for important sublayers and more for less important ones. Figure \ref{Cosine similarity} illustrates the cosine similarity among various sublayers. Based on the similarity metric, we develop a novel allocation strategy to determine appropriate compression ratios.

\begin{figure}[h]
  \centering
    \includegraphics[width=\columnwidth]{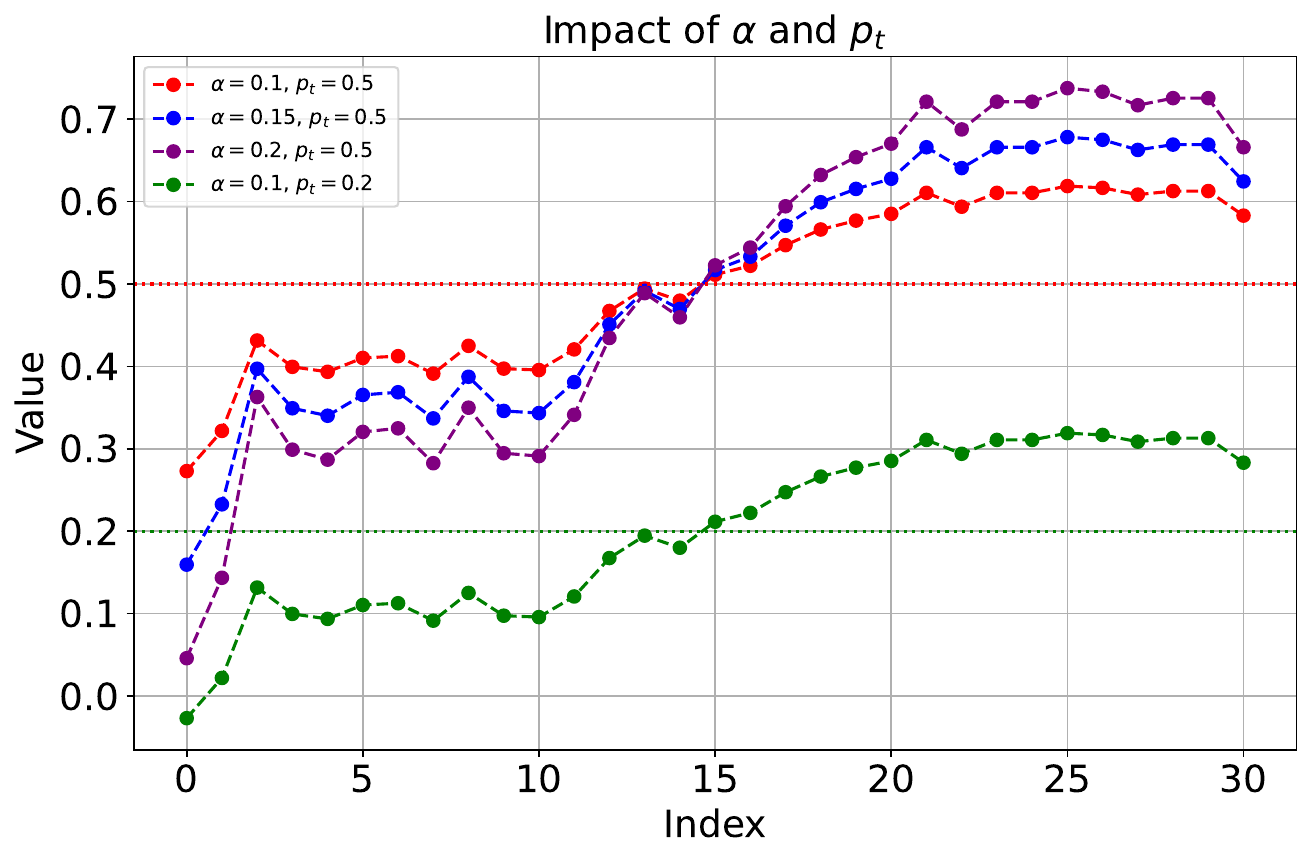}
  \caption{The influence of parameters $\alpha$ and $p_{t}$ on the normalized output $p$.}
  \label{fig:hyper}
\end{figure}

To effectively transform similarity measures into compression ratios, we propose a systematic allocation strategy that maintains the relative importance of different sublayers while satisfying the overall compression requirements. Our approach consists of three key steps: \emph{1)}  We apply $Z$-score normalization to standardize the distribution of cosine similarities across sublayers:
\begin{equation}
    z = \frac{I-\mathbb{E}[I]}{\sqrt{\mathrm{Var}[I]}}.
\end{equation}
\emph{2)} A scaling coefficient $\alpha$ is introduced to modulate the sensitivity of compression ratio. This hyperparameter governs the magnitude of sublayer compression variations,  where larger values of $\alpha$ induce more pronounced compression ratio differences.
\emph{3)} To satisfy the global compression constraint characterized by target ratio $p_t$, we apply a uniform translation to the normalized distribution:
\begin{align}\label{eq:allocate}
p = \alpha \cdot z + p_t = \alpha \cdot (\frac{I-\mathbb{E}[I]}{\sqrt{\mathrm{Var}[I]}}) + p_t,
\end{align}
where $\mathbb{E}[\cdot]$ and $\mathrm{Var}[\cdot]$ denote the expectation and variance operators respectively. As illustrated in Figure \ref{fig:hyper}, $\alpha$ and $p_t$ jointly determine the final compression ratio, with $\alpha$ controlling the relative compression differences between sublayers while $p_t$ ensures the overall compression target is met.

After establishing the general allocation framework, we further consider the architectural characteristics of transformer models. 
In Eq.(\ref{eq:allocate}), we allocate compression ratios to different sublayers based on their importance $I$, while ignoring the difference in the parameters quantity between different sublayers. The direct application of  Eq.(\ref{eq:allocate}) would lead to a mismatch between the target and actual compression ratios. For instance, in the LLaMA1 architecture, FFN sublayers contain twice the number of parameters as MHA sublayers. Assume that  $p_{t}$ is set as 0.5 and  the compression ratios of FFN sublayer $p_{ffn}$ and MHA sublayer $p_{mha}$ calculated by Eq.(\ref{eq:allocate}) are 0.4 and 0.6 respectively, the realized overall compression ratio becomes (0.6 + 0.4 × 2)/3 = 0.467, deviating from the target ratio of 0.5. To address this, we introduce $\beta$ as the ratio of parameters quantity  between FFN and MHA sublayers (e.g., $\beta=2$ in LLaMA1). Therefore, the actual  compression ratio $p_{s}$ is calculated as follows:
\begin{align}\label{ratio}
 p_{s} =\frac{(\mathbb{E}[p_{ffn}]*\beta+\mathbb{E}[p_{mha}])}{(\beta+1)}. 
\end{align}
We apply a one-time adjustment to the compression parameter $p$ through Eq. (\ref{final_ratio}). This  adjustment ensures that the final compression ratio precisely matches target $p_t$:
\begin{align}\label{final_ratio}
 p = p + (p_{t} - p_{s}).
\end{align}

\subsection{Energy-Balanced Matrix Parameter Allocation}
Within a sublayer, there typically exist multiple weight matrices, such as $\mathbf{W}_{q},\mathbf{W}_{k},\mathbf{W}_{v},\mathbf{W}_{o}$ in MHA sublayer. These weight matrices exhibit distinct eigenvalue distributions, which significantly impacts model compression. To investigate this phenomenon, we conducted preliminary experiments by individually compressing each weight matrix to 1024 ranks and evaluating the compressed model performance across multiple datasets. The experimental results in Table \ref{exploration energy} under the ``Fixed Rank" method reveal substantial performance variations when applying the same rank compression across different matrices. This observation strongly indicates that adopting the same retained rank across all weight matrices is suboptimal for model compression. Given this heterogeneous nature of weight matrix, the challenge lies in determining an optimal parameter allocation strategy for weight matrices. 

\renewcommand{\arraystretch}{1.1}
\begin{table}[htbp]
\centering
\setlength\tabcolsep{10pt}
\caption{The exploration of energy.}
\begin{tabular}{c|c|ccc}
\hline
Method & Matrix &  WikiText2 $\downarrow$ & PTB $\downarrow$  & C4 $\downarrow$ \\
\hline
\multirow{4}{*}{Fixed Rank}
 & $\mathbf{W}_{q}$ & 5.75 & 9.14 & 7.38 \\
 & $\mathbf{W}_{k}$ & 5.75 & 8.93 & 7.38 \\
 & $\mathbf{W}_{v}$ & 7.18 & 12.27 & 9.96 \\
 & $\mathbf{W}_{o}$ & 6.13 & 10.18 & 8.47 \\
\hline
\multirow{4}{*}{Fixed Energy}
 & $\mathbf{W}_{q}$ & 5.74 & 8.92 & 7.33 \\
 & $\mathbf{W}_{k}$ & 5.79 & 8.93 & 7.37 \\
 & $\mathbf{W}_{v}$ & 5.81 & 9.03 & 7.41 \\
 & $\mathbf{W}_{o}$ & 5.73 & 8.92 & 7.32 \\
\hline
\end{tabular}
\label{exploration energy}
\end{table}

 \begin{figure*}[h]
  \centering
    \subfigure{\includegraphics[width=0.45\textwidth]{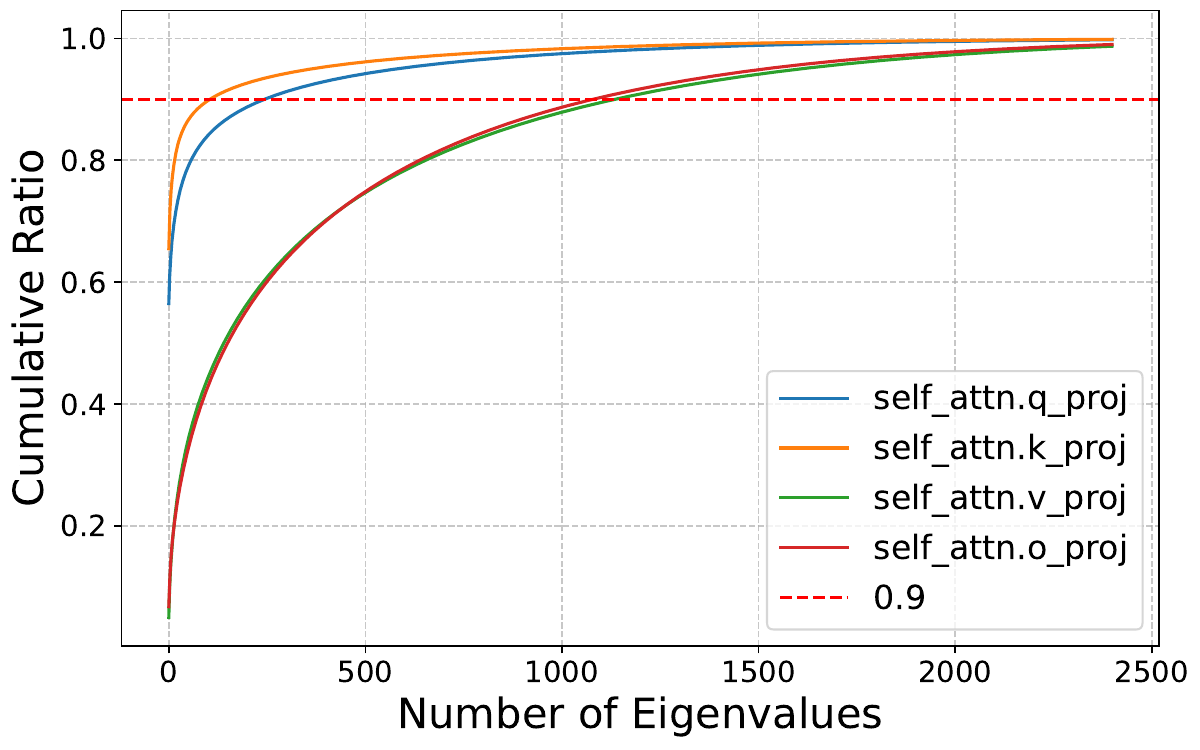}}
    \hspace{5pt} 
    \subfigure{\includegraphics[width=0.45\textwidth]{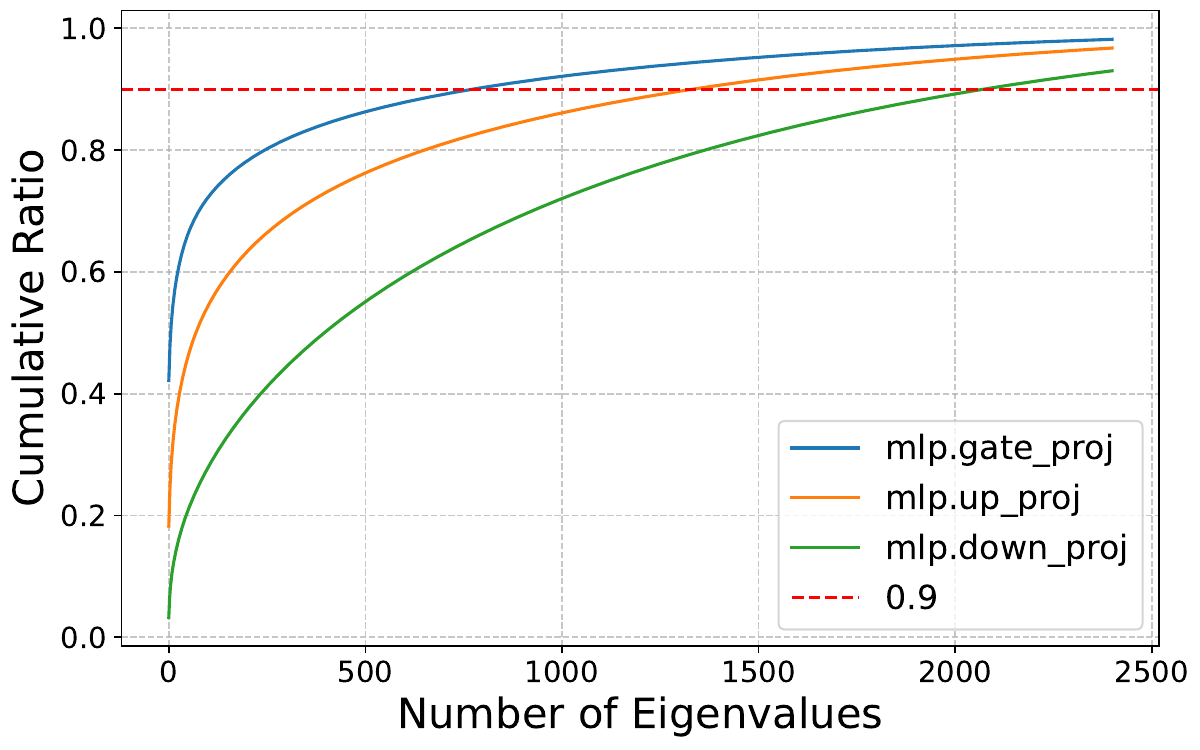}}
  \caption{The cumulative energy vectors of the four weight matrices in the MHA sublayer are shown in the left and the cumulative energy vectors of the three weight matrices in the FFN sublayer are shown in the right.}
  \label{MHA energy}
\end{figure*}
To address this challenge, we shall analyze it from an energy perspective. According to the PCA method involved in Eq. (\ref{autocorrelation matrix}), each eigenvalue $\lambda_{j}$ can be naturally interpreted as the energy, as it represents the variance of feature along its corresponding eigenvector direction. Since the absolute energy scales may vary across different matrices, examining their relative energy provides a more meaningful way to compare their importance. Therefore, to quantify the energy proportion of each eigenvalue in the matrix information, we propose a normalization strategy that scales each eigenvalue relative to the total energy of the matrix. This normalization enables direct comparison of energy distributions across different matrices:
\begin{equation}\label{single energy}
 \hat{\lambda}_{j} = \frac{\lambda_{j}}{\sum_{i=1}^{d_{out}} \lambda_i}.    
\end{equation}
After normalization, we calculate the cumulative energy vector. The cumulative energy vector $\mathbf{c}$ is constructed by sequentially summing the normalized eigenvalues:
\begin{align}\label{C}
\mathbf{c} = [c_1, \ldots, c_{d_{out}}] = \left[ \sum_{j=1}^{1} \hat{\lambda}_{j}, \sum_{j=1}^{2} \hat{\lambda}_{j}, \ldots, \sum_{j=1}^{d_{out}} \hat{\lambda}_{j} \right] .
\end{align}
Each element $c_i$ of the cumulative energy vector $\mathbf{c}$ represents the proportion of total matrix energy captured by the top $i$ eigenvalues, providing a comprehensive view of how energy is distributed.  
To investigate whether preserved energy ratio is a reliable indicator of model performance, we analyze both the energy distribution patterns and compression results under the ``Fixed Energy" method, as shown in Table \ref{exploration energy}. The experimental results demonstrate that when compressing different matrices while maintaining consistent energy ratios, the model performance remains notably comparable. This observation reveals a strong correlation between preserved energy ratio and model performance. Further analysis of the energy distribution patterns, as shown in Figure \ref{MHA energy}, indicates that weight matrices within both MHA and FFN sublayers exhibit distinct energy distribution characteristics, with different projection matrices requiring varying ranks to preserve the same proportion of total energy. This finding suggests that we can leverage these energy distribution differences to allocate different compression ratios to different weight matrices, as long as we ensure consistent preserved energy ratios.

\begin{algorithm}[!h]
\caption{MGAA}
\label{mgaa algorithm}
\textbf{Input}: Uncompressed sublayer set $\mathcal{M}$; Target compression ratio $p_t$; Scaling factor $\alpha$; \\
\textbf{Output}: Compressed set $\mathcal{M}'$

\begin{algorithmic}[1]
\STATE Initialize $\mathcal{M}' \leftarrow \{\}$
\FOR{sublayer $\mathcal{S}$ in $\mathcal{M}$}
    \STATE Compute importance $I$ by cosine similarity Eq. \eqref{importance}
\ENDFOR

\STATE Compute compression ratios $p$ by Eq. \eqref{eq:allocate}, \eqref{ratio} ,\eqref{final_ratio}
\FOR{sublayer $\mathcal{S}$ in $\mathcal{M}$}
    \STATE $\mathcal{W} \leftarrow$ weight matrices in $\mathcal{S}$
    \FOR{$\mathbf{W}$ in $\mathcal{W}$}
        \STATE Compute cumulative energy vector $\mathbf{c}$ by Eq. \eqref{C}
    \ENDFOR
    \STATE Compute retained rank by Eq. \eqref{energy limit}
    \STATE Compress matrices by Eq. \eqref{compression}
\ENDFOR
\end{algorithmic}
\end{algorithm}

Based on these, we formulate the matrix compression problem as a constrained optimization problem:
\begin{equation}\label{energy limit}
\begin{split}
&\max \sum_{a\in T} c_{r_a} \\ 
& \text{s.t. }  \sum_{a \in \mathcal{T}} r_a \leq R_{\text{budget}} \\ 
& \forall_{a,b \in \mathcal{T}} \;  \left|c_{r_a} - c_{r_b}\right| < \epsilon
\end{split}
\end{equation}
where $\mathcal{T}$ denotes the set of weight matrices in the sublayer, $r_a$ represents the allocated rank for matrix $\mathbf{W}_{a}$, $c_{r_a}$ is the retained energy ratio when matrix $\mathbf{W}_{a}$ is compressed to rank $r_a$, $R_{\text{budget}}$ defines the total rank budget for the sublayer which is determined by the sublayer compression ratio. The objective is to maximize the sum of preserved energy ratios across all weight matrices under a given computational budget constraint. To maintain consistent model performance, we introduce an energy balance constraint, which ensures that the difference in preserved energy ratios between any two matrices does not exceed a threshold $\epsilon$. This problem is a typical convex optimization problem and can be solved using various methods such as binary search, dynamic programming, or mixed-integer programming. Additionally, we establish a lower bound on the parameter retention ratio for weight matrix to prevent excessive parameter reduction in any single matrix. This lower bound serves as a safeguard mechanism, ensuring a balanced compression across different matrices and maintaining the model's structural integrity. We outline the  MGAA algorithm in Algorithm \ref{mgaa algorithm}.

\subsection{Discussion}
In this subsection, we present a comprehensive discussion delineating three principal advantages that distinguish our methodological approach from previous studies.

\indent \emph{1) Computational Efficiency.}
Our approach offers substantial computational advantages by leveraging only activation values from the forward pass, thus circumventing the computational burden of gradient calculations \cite{FWSVD} in large models. A key strength of our method lies in its streamlined compression, which can be executed in merely two forward propagation steps. This stands in marked contrast to conventional heuristic approaches \cite{AFM} that necessitate multiple iterative adjustments based on downstream validation performance. By eliminating these recursive optimization steps, our method achieves remarkable efficiency, significantly reducing the compression time from several hours to just a few minutes.

\indent \emph{2) Generalization Ability.}
The heuristic compression methods \cite{AFM} allocate compression ratios based on the performance on specific downstream tasks, which may lead to overfitting to the characteristics of the target dataset. In contrast, our approach adopts task-agnostic, general-purpose data as calibration data, thereby reducing the risk of overfitting and ensuring that the compressed model retains its generalization ability across diverse tasks. This design not only enhances the robustness of the compressed model but also demonstrates its adaptability to unseen tasks, as evidenced by the improved accuracy in zero-shot learning scenarios in the following Section \ref{section:experiments}. Such a strategy highlights the superiority of our method in maintaining a balance between compression efficiency and model versatility, making it more suitable for real-world applications where task-specific data may be limited or unavailable.

\indent \emph{3) Plug-and-play Module.}
While  MGAA was primarily developed for the feature space compression paradigm described in Eq. (\ref{layer-wise}), it can be extended to the weighted matrix approximation paradigm as shown in Eq. (\ref{layer-wise-question}).  These two paradigms differ in their operational domains and energy distribution representations: the feature space compression paradigm operates on the autocorrelation matrix, where eigenvalues directly represent the energy distribution along principal components. In contrast, the weighted matrix approximation framework applies SVD directly to the weighted matrices, where the energy distribution is represented by the squared singular values $\sigma^2$ along their respective eigenvector directions. The specific cumulative energy vector $\mathbf{c}$ is calculated as follows:
\begin{equation}\label{svd_energy}
\begin{split}
&\mathbf{\Sigma}=\text{diag}(\sigma_1,\sigma_2,\ldots,\sigma_{d_{out}}), \\
  &\hat{\sigma}_{j} = \frac{\sigma_{j}^{2}}{\sum_{i=1}^{d_{out}} \sigma_i^{2}},  \\
  & \mathbf{c} =  \left[ \sum_{j=1}^{1} \hat{\sigma}_{j}, \sum_{j=1}^{2} \hat{\sigma}_{j}, \ldots, \sum_{j=1}^{d_{out}} \hat{\sigma}_{j} \right] .
\end{split}
\end{equation}
The ability of MGAA to handle both paradigms demonstrates its broad applicability in low-rank approximation frameworks.

\section{Experiments} \label{section:experiments}
\subsection{Experimental Settings}
In this subsection, we elaborate the backbone, calibration datasets, model evaluation and baseline methods.

\emph{1) The Backbones.} We evaluate our approach on various state-of-the-art LLMs, including LLaMA1, Vicuna, LLaMA3, and Mistral, to demonstrate its effectiveness and generalization capability across different model architectures. \\
\indent \emph{2) Calibration Datasets.}
The performance of the compressed model is related to the calibration data. A comprehensive analysis of the calibration data will be presented in ablation study. To preserve the language modeling and common sense reasoning inherent to LLMs, we carefully select calibration data that represents diverse linguistic patterns and reasoning tasks. Specifically, we utilize: 1) WikiText2 \cite{WikiText2}: A dataset of high-quality Wikipedia articles that captures general language patterns, 2) Alpaca \cite{alpaca}: A dataset focusing on instruction-following. From each dataset, we sample 128 sequences, each containing 512 tokens, ensuring a balanced and diverse calibration data. \\
\indent \emph{3) Model Evaluation.}
The comprehensive evaluation of models' zero-shot performance is conducted along two critical aspects: language modeling and common sense reasoning. To assess language modeling capability, we evaluate the perplexity of the compressed model on the WikiText2 dataset. Each test sequence contains 2048 tokens. And we use accuracy to evaluate the reasoning capability on seven datasets:
1) \textbf{BoolQ} \cite{BoolQ}: A natural language understanding dataset of 14.5K yes/no questions testing factual information extraction.
2) \textbf{PIQA} \cite{PIQA}: A physical commonsense dataset containing 21K questions focusing on practical reasoning and problem-solving.
3) \textbf{HellaSwag} \cite{HellaSwag}: A dataset of 40K multiple-choice questions testing commonsense inference and contextual understanding.
4) \textbf{WinoGrande} \cite{WinoGrande}: A challenging dataset of 44K pronoun resolution problems for evaluating commonsense reasoning.
5) \textbf{ARC-easy} \cite{ARC-easy}: A collection of 6K elementary science questions focusing on basic knowledge and reasoning.
6) \textbf{ARC-challenge} \cite{ARC-easy}: A more challenging version of ARC testing complex scientific reasoning and comprehension.
7) \textbf{OpenbookQA} \cite{OPENQA}: A dataset of 5.9K questions requiring both factual knowledge and reasoning ability. All evaluations are conducted using the lm-evaluation-harness framework \cite{lm-eval} to ensure consistency and reproducibility. \\
\indent \emph{4) Baseline Methods.} We compare our method against several low-rank approximation approaches:
\begin{itemize}
    \item SVD \cite{first_svd}: A fundamental matrix factorization method that applies truncated SVD directly to the weight matrix, without relying on calibration data.

    \item Activation-aware SVD (ASVD)\cite{ASVD}: An advanced approach that leverages the mean absolute value of input activations as a weighting mechanism prior to SVD decomposition. The final approximation is obtained by incorporating the inverse weights into the low-rank matrix derived from the weighted matrix's truncated SVD.

    \item Activation-weighted SVD (AWSVD)\cite{lorap}: This method computes the Euclidean norm ($\ell_2$ norm) of input activations and employs these values for matrix weighting. Steps including  truncated SVD and inverse weighting, are consistent with ASVD.

    \item Principal Components Analysis (PCA)\cite{SVD-llm}: This method performs EVD on the output features' autocorrelation matrix, and then project the weight matrix onto the principal components of the resulting orthogonal matrix to obtain low-rank matrices.
    
    \item  Atomic Feature Mimicking (AFM)\cite{AFM}: This method performs EVD on the output features' covariance matrix, disregarding activation scales information. The subsequent computational steps are identical to PCA.

\end{itemize}


\subsection{Experimental Results}
\indent \emph{1) Language Modeling.}
We evaluated the performance of six low-rank approximation methods at 20\% and 50\% compression ratios. The results are presented in Table \ref{main-result}, where lower perplexity values indicate better performance. Our proposed method MGAA demonstrates superior performance across all models compared to other low-rank approximation techniques. This superiority becomes more pronounced as the compression ratio increases. At 20\% compression ratio, MGAA achieves consistent perplexity reductions across all models. The improvements become more substantial at 50\% compression ratio, where MGAA significantly outperforms the strongest baseline method (PCA). Most notably, when compressing LLaMA1, MGAA achieves a perplexity of 22.02 compared to PCA's 33.21, representing a significant 33.7\% improvement.  On average, across all four models at 50\% compression, MGAA demonstrates an 18\% reduction in perplexity compared to PCA.
Among the baseline methods, PCA stands out for its unique ability to maintain stable performance even at the high compression ratio of 50\%. This robustness can be attributed to PCA's fundamental property of directly relating compression loss to the sum of truncated eigenvalues, enabling effective preservation of model performance under aggressive compression.

\indent \emph{2) Reasoning Ability.}
Table \ref{main-result} presents the average accuracy of compressed models across seven common sense reasoning datasets at 20\% and 50\% compression ratios. Our proposed MGAA method demonstrates superior performance across various compression ratios and language models. At a compression ratio of 20\%, MGAA outperforms other methods on three models (LLaMA1, Vicuna, 
 and LLaMA3). For the Mistral model, weighted decomposition methods ASVD and AWSVD show slightly better performance. Notably, MGAA maintains its efficacy even at a high compression ratio of 50\%, consistently surpassing other methods by at least 3\% in accuracy. As compression ratio increases from 20\% to 50\%, the performance  decline is less pronounced for MGAA compared to other methods, particularly for models like LLaMA1 and Vicuna. These findings underscore MGAA's effectiveness and robustness as a compression approach, demonstrated by its consistent superior performance across diverse model architectures and compression ratios.

\renewcommand{\arraystretch}{1.1}
\begin{table*}[!h]
\vspace*{0pt} 
\caption{The  results  on multiple backbones. At the same compression ratio, \textbf{`bold'} represents the best performance.}
  \centering
   \setlength\tabcolsep{3.2pt}
  \scriptsize
   \begin{threeparttable}
  \begin{tabular}{c|cc|ccc|cccccccc}
    \toprule
    \toprule
   Model & Compression Ratio & Method  & WikiText2$\downarrow$ & PTB$\downarrow$ & C4$\downarrow$ &BoolQ$\uparrow$ & PIQA$\uparrow$ & HellaSwag$\uparrow$ & WinoGrande$\uparrow$ & ARC-e$\uparrow$ & ARC-c$\uparrow$ & OBQA$\uparrow$ & Average$\uparrow$ \\
   \midrule

   LLaMA1 & Ratio=0\% & Baseline & 5.68 & 8.80 & 7.24 & 74.98 & 78.67 & 70.00 & 76.21 & 75.34 & 44.62 & 44.40 & 66.32 \\
   \midrule
\multirow{6}{*}{\shortstack{LLaMA1}} & \multirow{6}{*}{\shortstack{Ratio=20\%}}
                                & SVD & $2.00E^{4}$ & $2.02E^{4}$ & $1.75E^{4}$ & 41.22 & 53.86 & 51.14 & 27.14 & 27.31 & 24.91 & 26.20 & 35.97 \\
                            &   & ASVD   & 22.68 & 32.75 & 26.98 & 66.64 & 65.61 & 63.46 & 52.22 & 54.97 & 30.72 & 35.60 & 52.75 \\
                            &   & AWSVD   & 17.61 & 27.42 & 22.63 & \textbf{70.28} & 68.01 & 63.61 & 56.03 & 57.87 & 33.70 & 38.60 & 55.44 \\
                            &   & AFM   & 21.42 & 60.82 & 40.80 & 63.18 & 65.83 & 59.75 & 51.90 & 52.86 & 31.91 & 34.80 & 51.46 \\
                            &   & PCA  & 8.75 & 15.17 & 14.17 & 63.52 & 72.96 & 67.80 & 63.19 & 66.84 & 36.18 & 38.60 & 58.44 \\
                            &   & MGAA   & \textbf{7.65} & \textbf{12.15} & \textbf{11.37} & 62.32 & \textbf{74.27} & \textbf{68.51} & \textbf{68.39} & \textbf{69.19} & \textbf{38.23} & \textbf{41.60} & \textbf{60.36} \\                        
    \midrule
    \multirow{6}{*}{\shortstack{LLaMA1}} & \multirow{6}{*}{\shortstack{Ratio=50\%}}
                                & SVD & $1.31 E^{5}$ & $9.00 E^{4}$ & $7.96 E^{4}$ & \textbf{59.02} & 52.07 & 50.43 & 25.98 & 26.60 & \textbf{28.84} & 26.40 & 38.48 \\
                            &   & ASVD   & $1.07 E^{3}$ & $1.23 E^{3}$ & $1.26 E^{3}$ & 37.83 & 53.05 & 49.80 & 27.35 & 27.23 & 26.37 & 24.40 & 35.15 \\
                            &   & AWSVD   & 705.57 & 765.67 & 668.98 & 42.17 & 54.24 & 49.80 & 28.00 & 27.65 & 25.68 & 22.00 & 35.65 \\
                            &   & AFM   & $3.37 E^{3}$ & $3.81 E^{3}$ & $3.59 E^{3}$ & 45.20 & 53.43 & 51.70 & 27.32 & 29.80 & 23.98 & 27.20 & 36.95 \\
                            &   & PCA  & 33.21 & 80.07 & 68.76 & 54.04 & 58.00 & 54.62 & 35.58 & 38.72 & 26.37 & 29.00 & 42.33 \\
                            &   & MGAA   & \textbf{22.02} & 58.57 & \textbf{43.33} & \textbf{57.55} & \textbf{63.87} & \textbf{60.06} & \textbf{41.82} & \textbf{47.64} & 27.90 & \textbf{34.00} & \textbf{47.55} \\  
    \midrule
    \midrule
   Vicuna & Ratio=0\% & Baseline & 6.78 & 26.78 & 9.08 & 76.92 & 77.26 & 69.38 & 73.77 & 75.63 & 45.73 & 45.00 & 66.24 \\
   \midrule
\multirow{6}{*}{\shortstack{Vicuna}} & \multirow{6}{*}{\shortstack{Ratio=20\%}}
                                & SVD & $1.89 E^{4}$ & $2.74 E^{4}$ & $1.57 E^{4}$ & 48.20 & 52.12 & 49.49 & 26.45 & 25.59 & 27.82 & 27.00 & 36.67 \\
                            &   & ASVD   & 57.92 & 304.53 & 74.61 & 63.12 & 64.15 & 55.17 & 42.43 & 48.57 & 28.75 & 33.20 & 47.91 \\
                            &   & AWSVD   & 45.90 & 285.77 & 59.35 & 64.50 & 64.04 & 57.85 & 44.40 & 51.14 & 29.01 & 31.40 & 48.91 \\
                            &   & AFM   & 58.79 & 241.14 & 70.87 & 60.98 & 60.07 & 52.80 & 38.83 & 45.03 & 28.92 & 30.00 & 45.23 \\
                            &   & PCA   & 10.63 & 51.33 & 17.99 & \textbf{71.13} & 70.13 & 63.54 & 57.70 & 65.78 & 38.23 & 38.60 & 57.87 \\
                            &   & MGAA   & \textbf{8.86} & \textbf{36.31} & \textbf{13.42} & 67.95 & \textbf{73.88} & \textbf{66.14} & \textbf{65.16} & \textbf{71.34} & \textbf{40.02} & \textbf{42.00} & \textbf{60.93} \\
                       
    \midrule
    \multirow{6}{*}{\shortstack{Vicuna}} & \multirow{6}{*}{\shortstack{Ratio=50\%}}
                                & SVD & $3.79 E^{4}$ & $3.96 E^{4}$ & $3.80 E^{4}$ & 40.80 & 52.45 & 50.12 & 26.09 & 25.08 & 28.07 & 24.80 & 35.34 \\
                            &   & ASVD   & $9.37 E^{2}$ & $2.52 E^{3}$ & $9.40 E^{2}$ & 54.86 & 52.50 & 49.01 & 26.93 & 26.35 & 26.02 & 25.20 & 37.27 \\
                            &   & AWSVD   & 800.94 & $2.01 E^{3}$ & 735.07 & 50.98 & 54.19 & 50.12 & 27.75 & 27.48 & 25.94 & 25.60 & 37.44 \\
                            &   & AFM  & $4.22 E^{3}$ & $3.47 E^{4}$ & $2.02 E^{3}$ & 44.92 & 54.73 & 48.62 & 26.68 & 26.30 & 24.57 & 26.40 & 36.03 \\
                            &   & PCA   & 37.72 & 272.34 & 79.47 & 46.51 & 58.32 & 54.06 & 33.59 & 39.39 & 26.19 & 29.00 & 41.01 \\
                            &   & MGAA   & \textbf{28.33} & \textbf{195.98} & \textbf{58.56} & \textbf{58.93} & \textbf{61.32} & \textbf{56.35} & \textbf{37.33} & \textbf{45.08} & \textbf{26.62} & \textbf{30.60} & \textbf{45.18} \\
    \midrule
    \midrule
 LLaMA3 & Ratio=0\% & Baseline & 6.14 & 10.58 & 9.31 & 81.31 & 79.71 & 72.61 & 79.15 & 80.09 & 53.41 & 45.00 & 70.18 \\
   \midrule
\multirow{6}{*}{\shortstack{LLaMA3}} & \multirow{6}{*}{\shortstack{Ratio=20\%}}
                                & SVD & $6.27 E^{5}$ & $3.06 E^{5}$ & $4.61 E^{5}$ & 38.56 & 53.05 & 50.67 & 26.71 & 26.09 & 26.37 & 29.80 & 35.89 \\
                            &   & ASVD  & 193.33 & 637.46 & 225.55 & 65.47 & 62.68 & 56.43 & 40.44 & 44.32 & 27.39 & 30.00 & 46.68 \\
                            &   & AWSVD   & 171.83 & 497.11 & 197.57 & \textbf{66.09} & 63.66 & 56.75 & 42.71 & 47.69 & 28.24 & 31.20 & 48.05 \\
                            &   & AFM  & 151.15 & 494.02 & 425.51 & 40.31 & 60.23 & 54.14 & 36.33 & 40.24 & 26.11 & 30.40 & 41.11 \\
                            &   & PCA   & 18.47 & 49.86 & 35.91 & 64.62 & 70.08 & 66.93 & 56.47 & 63.09 & 36.01 & 38.00 & 56.46 \\
                            &   & MGAA   & \textbf{11.57} & \textbf{23.63} & \textbf{20.89} & 63.18 & \textbf{72.96} & \textbf{69.53} & \textbf{63.80} & \textbf{69.74} & \textbf{42.06} & \textbf{40.60} & \textbf{60.27} \\                       
    \midrule
    \multirow{6}{*}{\shortstack{LLaMA3}} & \multirow{6}{*}{\shortstack{Ratio=50\%}}
                                & SVD & $1.36 E^{5}$ & $1.07 E^{5}$ & $1.42 E^{5}$ & 44.65 & 53.59 & 51.62 & 26.24 & 24.16 & 25.94 & 30.00 & 36.60 \\
                            &   & ASVD   & $1.53 E^{4}$ & $1.99 E^{4}$ & $1.25 E^{4}$ & 44.50 & 52.61 & 50.28 & 25.97 & 24.75 & 26.02 & \textbf{30.80} & 36.42 \\
                            &   & AWSVD   & $1.69 E^{4}$ & $2.07 E^{4}$ & $1.12 E^{4}$ & 38.29 & 53.92 & 49.88 & 25.92 & 26.18 & \textbf{27.65} & 29.20 & 35.86 \\
                            &   & AFM   & 835.30 & $1.43 E^{3}$ & $2.18 E^{3}$ & 37.77 & 54.46 & 50.59 & 27.29 & 27.10 & 23.63 & 26.20 & 35.29 \\
                            &   & PCA   & 97.64 & 407.53 & 218.18 & 39.11 & 55.66 & 50.28 & 29.12 & 31.02 & 22.87 & 25.60 & 36.24 \\
                            &   & MGAA   & \textbf{64.76} & \textbf{349.12} & \textbf{135.28} & \textbf{61.59} & \textbf{59.74} & \textbf{54.62} & \textbf{32.37} & \textbf{37.21} & 23.72 & 25.40 & \textbf{42.09 }\\  
    \midrule
    \midrule
     Mistral & Ratio=0\% & Baseline & 5.32 & 23.00 & 8.26 & 82.05 & 80.14 & 73.80 & 80.43 & 79.55 & 52.30 & 44.20 & 70.35 \\
   \midrule
\multirow{6}{*}{\shortstack{Mistral}} & \multirow{6}{*}{\shortstack{Ratio=20\%}}
                                & SVD & $7.53 E^{5}$ & $7.23 E^{4}$ & $6.29 E^{5}$ & 56.82 & 52.83 & 50.59 & 25.88 & 25.72 & 28.67 & 26.00 & 38.07 \\
                            &   & ASVD   & 13.53 & 65.94 & 18.07 & 79.24 & 74.16 & 67.25 & 63.02 & 68.22 & 37.97 & 35.60 & 60.78 \\
                            &   & AWSVD   & 12.90 & 63.89 & 17.64 & \textbf{79.66} & 73.50 & 67.40 & 63.87 & 68.18 & 38.91 & 38.20 & 61.39 \\
                            &   & AFM   & 10.71 & 88.77 & 20.92 & 64.07 & 71.16 & 65.35 & 58.33 & 64.48 & 35.49 & 35.40 & 56.33 \\
                            &   & PCA   & 9.26 & 55.14 & 18.63 & 65.99 & 72.03 & 66.77 & 60.43 & 67.38 & 36.77 & 36.20 & 57.94 \\
                            &   & MGAA   & \textbf{7.54} & \textbf{40.82} & \textbf{13.66} & 66.36 & \textbf{74.59} & \textbf{67.96} & \textbf{65.91} & \textbf{69.57} & \textbf{39.59} & \textbf{38.40} & \textbf{60.34} \\                       
    \midrule
    \multirow{6}{*}{\shortstack{Mistral}} & \multirow{6}{*}{\shortstack{Ratio=50\%}}
                                & SVD & $3.87 E^{4}$ & $3.43 E^{4}$ & $3.69 E^{4}$ & 46.45 & 52.83 & 50.28 & 25.54 & 26.09 & \textbf{29.01} & 25.80 & 36.57 \\
                            &   & ASVD  & 241.95 & 571.58 & 234.64 & 41.13 & 52.45 & 50.12 & 29.17 & 30.09 & 23.29 & 23.80 & 35.72 \\
                            &   & AWSVD   & 184.36 & 445.18 & 177.99 & 44.83 & 53.59 & 49.01 & 29.61 & 30.05 & 23.12 & 24.20 & 36.34 \\
                            &   & AFM   & 55.86 & 228.10 & 109.04 & 39.85 & 55.50 & 52.33 & 31.11 & 32.66 & 22.78 & 27.20 & 37.35 \\
                            &   & PCA   & 36.79 & 227.79 & 92.00 & 40.18 & 56.42 & 53.35 & 31.92 & 33.29 & 24.23 & 24.20 & 37.66 \\
                            &   & MGAA   & \textbf{31.34} & \textbf{204.27} & \textbf{71.39} & \textbf{46.76} & \textbf{59.30} & \textbf{55.17} & \textbf{34.48} & \textbf{39.27} & 25.77 & \textbf{27.60} & \textbf{41.19} \\

    \midrule
    \bottomrule
  \end{tabular}
    \end{threeparttable}
      \label{main-result}
\end{table*}


\subsection{Experiments on Larger Model}
To further validate the scalability and effectiveness of MGAA on larger models, we conducted extensive experiments on the LLaMA1-13B model with more aggressive compression ratios (40\%-70\%) compared to our previous experiments on 7B architectures. The results in Table \ref{13B} demonstrate MGAA's consistent superiority over the baseline PCA method across all compression scenarios. At 40\% compression ratio, MGAA reduces perplexity from 13.46, 25.04, and 25.05 to 10.53, 19.23, and 18.65 on WikiText2, PTB, and C4 respectively, while improving the average accuracy on downstream tasks by 3.2 percentage points (from 52.40\% to 55.60\%). Even under aggressive 70\% compression, MGAA maintains its significant advantage, achieving a 22.5\% perplexity reduction on WikiText2 (from 76.25 to 52.37) while exhibiting minimal performance degradation on downstream tasks. The enhanced performance gains observed in scaling from 7B to 13B parameters further validate MGAA's potential for larger model. These comprehensive results demonstrate MGAA as a robust and efficient plug-and-play compression method, particularly effective for large-scale models under high compression ratio.

\subsection{Ablation Study and More In-Depth Analysis}
\indent \emph{1) Effectiveness of MGAA Components.}
To evaluate our proposed sublayer-wise compression ratio allocation and energy-balanced parameter allocation methods, we independently apply the sublayer-wise compression ratio allocation and energy-balanced parameter allocation to the PCA method, creating two variants: L-PCA (Layer-wise PCA) and E-PCA (Energy-balanced PCA). The experimental results on LLaMA1 are shown in Table \ref{L-PCA}. The experimental findings demonstrate that both of our proposed components consistently enhance the performance of the compressed model across various compression ratios. The advantages of L-PCA and E-PCA over the original PCA become more substantial as the compression ratio increases, with improvements observed in both language modeling perplexity and downstream task accuracy. 
Furthermore, the results suggest that the L-PCA and E-PCA methods demonstrate complementary benefits, and can be combined to achieve even greater performance improvements. These results highlight the effectiveness of  multi-granular adaptive allocation strategies in optimizing the compression of large language models. 

\renewcommand{\arraystretch}{1.1}
\begin{table}[htbp]
\centering
\setlength\tabcolsep{6.5pt}
\caption{The experiment results with diverse compression rations on LLaMA1-13B.}
\begin{tabular}{c|c|ccc|c}
\hline
Method & Ratio & WikiText2 $\downarrow$ & PTB $\downarrow$  & C4 $\downarrow$ & Average $\uparrow$ \\
\hline
\multirow{4}{*}{PCA}
 & 40\% & 13.46 & 25.04 & 25.05 & 52.40 \\
 & 50\% & 22.15 & 44.66 & 44.56 & 46.27 \\
 & 60\% & 39.03 & 80.09 & 64.55 & 41.77 \\
 & 70\% & 76.25 & 165.82 & 160.68 & 36.66 \\
\hline
\multirow{4}{*}{+MGAA}
 & 40\% & 10.53 & 19.23 & 18.65 & 55.60 \\
 & 50\% & 16.20 & 33.06 & 31.21 & 50.32 \\
 & 60\% & 28.29 & 65.20 & 58.57 & 44.45 \\
 & 70\% & 52.37 & 129.74 & 112.39 & 39.94 \\
\hline
\end{tabular}
\label{13B}
\end{table}

\indent \emph{2) Application to Other Low-Rank Approximation Methods.}
MGAA is highly compatible and easily integrated with various low-rank approximation methods as a plug-and-play method, including ASVD \cite{ASVD}, AWSVD \cite{lorap}, AFM \cite{AFM}, and Joint Rank-k Approximation \cite{joint}, which performs low-rank approximation on row-wise concatenated query weight matrix $\mathbf{W_{q}}$ and key weight matrix $\mathbf{W_{k}}$. The integration requires careful consideration of the energy metrics used in different methods. Based on the previous analysis, when applying MGAA to weight decomposition methods (AWSVD \cite{lorap} and ASVD \cite{ASVD}), we adopt squared singular values as the energy metric to maintain theoretical consistency. To evaluate the effectiveness of this integration, we conducted comprehensive experiments combining MGAA with both feature-based (AFM \cite{AFM} ) and weight-based (AWSVD \cite{lorap}, ASVD \cite{ASVD}) decomposition methods, as well as Joint Rank-k Approximation. As shown in Table \ref{other_method}, MGAA consistently improves the performance of all baseline methods across compression ratios ranging from 20\% to 50\%. This improvements are consistent at higher compression ratios, demonstrating the robustness of our approach. These experiments not only validate the extensibility and adaptability of our method but also suggest promising directions for developing more efficient hybrid compression strategies.

\indent \emph{3) Affect of the Similarity Measures.}
While cosine similarity served as our primary metric, we also investigated the effectiveness of other distance measures, including Euclidean distance, Manhattan distance, and Normalized Euclidean distance. We present the experimental results under 50\% compression ratio in Table \ref{distance explor}. The results demonstrate that cosine similarity achieves superior performance across WikiText2, PTB, and C4 datasets (with scores of 24.54, 66.86, and 50.22 respectively) and attains the highest average accuracy (44.59) on inference datasets. These results indicate that cosine similarity is more effective at capturing internal feature transformations within the model, exhibiting significant advantages in similarity measurement tasks.

\begin{table}[htbp]
\centering
\setlength\tabcolsep{6pt}
\caption{Comparative analysis of different component configurations in MGAA.}
\begin{tabular}{c|c|ccc|c}
\hline
Method & Ratio & WikiText2 $\downarrow$ & PTB $\downarrow$  & C4 $\downarrow$ & Average $\uparrow$ \\
\hline
\multirow{4}{*}{PCA} 
 & 20\% & 8.75 & 15.17 & 14.17 & 58.44 \\
 & 30\% & 11.49 & 22.11 & 20.37 & 54.35 \\
 & 40\% & 18.17 & 39.94 & 34.73 & 49.17 \\
 & 50\% & 33.21 & 80.07 & 68.76 & 42.33 \\
\hline
\multirow{4}{*}{L-PCA} 
 & 20\% & 8.31 & 14.55 & 13.08 & 58.99 \\
 & 30\% & 10.63 & 20.71 & 18.36 & 55.45 \\
 & 40\% & 15.89 & 35.27 & 30.09 & 50.21 \\
 & 50\% & 24.54 & 66.86 & 50.22 & 44.59 \\
\hline
\multirow{4}{*}{E-PCA} 
 & 20\% & 8.15 & 12.86 & 12.61 & 59.02 \\
 & 30\% & 10.56 & 17.51 & 17.54 & 55.87 \\
 & 40\% & 16.68 & 35.12 & 31.10 & 50.30 \\
 & 50\% & 29.48 & 70.15 & 59.43 & 44.54 \\
\hline
\multirow{4}{*}{\parbox{1.5cm}{\quad PCA\\+MGAA}}
 & 20\% & 7.65 & 12.15  & 11.37 & 60.36  \\
 & 30\% & 9.61 & 17.18 & 15.99 & 57.12 \\
 & 40\% & 14.46 & 30.27 & 26.33 & 52.31 \\
 & 50\% & 22.02  & 58.57  & 43.33  & 47.55 \\
\hline
\end{tabular}
\label{L-PCA}
\end{table}

\renewcommand{\arraystretch}{1.2}
\begin{table}[htbp]
\centering
\setlength\tabcolsep{6pt}
\caption{The performance improvement after combining with other Low-rank Approximation method.}
\begin{tabular}{c|c|ccc|c}
\hline
Method & Ratio & WikiText2 $\downarrow$ & PTB $\downarrow$  & C4 $\downarrow$ & Average $\uparrow$ \\
\hline
\multirow{4}{*}{AFM}
 & 20\% & 21.42 & 60.82 & 40.80 & 51.46 \\
 & 30\% & 151.02 & 539.36 & 728.46 & 42.13 \\
 & 40\% & $1.71E^4$ & $2.56E^4$ & $2.20E^4$ &38.57 \\
 & 50\% & $3.37E^4$ & $3.81E^4$ & $3.59E^4$ & 36.95 \\
\hline
\multirow{4}{*}{+MGAA}
 & 20\% & 10.08 & 16.10 & 14.04 & 59.89 \\
 & 30\% & 14.70 & 27.16 & 21.97 & 54.83 \\
 & 40\% & 30.29 & 69.62 & 46.30 & 45.74 \\
 & 50\% & 63.09 & 199.11 & 106.48 & 39.79 \\
\hline
\multirow{4}{*}{ASVD}
 & 20\% & 22.68 & 32.75 & 26.98 & 52.75 \\
 & 30\% & 63.82 & 83.60 & 67.90 & 43.38 \\
 & 40\% & 205.98 & 278.39 & 206.90 &35.99 \\
 & 50\% & $1.07E^4$ & $1.23E^4$ & $1.26E^4$ & 35.15 \\
\hline
\multirow{4}{*}{+MGAA}
 & 20\% & 10.09 & 15.18 & 14.31 & 61.32 \\
 & 30\% & 15.91 & 23.15 & 23.21 & 56.39 \\
 & 40\% & 43.99 & 76.71 & 61.21 & 48.31 \\
 & 50\% & 182.34 & 268.94 & 235.36 & 41.85 \\
\hline
\multirow{4}{*}{AFM}
 & 20\% & 21.42 & 60.82 & 40.80 & 51.46 \\
 & 30\% & 151.02 & 539.36 & 728.46 & 42.13 \\
 & 40\% & $1.71E^4$ & $2.56E^4$ & $2.20E^4$ &38.57 \\
 & 50\% & $3.37E^4$ & $3.81E^4$ & $3.59E^4$ & 36.95 \\
\hline
\multirow{4}{*}{+MGAA}
 & 20\% & 10.08 & 16.10 & 14.04 & 59.89 \\
 & 30\% & 14.70 & 27.16 & 21.97 & 54.83 \\
 & 40\% & 30.29 & 69.62 & 46.30 & 45.74 \\
 & 50\% & 63.09 & 199.11 & 106.48 & 39.79 \\
\hline
\multirow{4}{*}{Joint}
 & 20\% & 8.48 & 15.00 & 13.78 & 58.35 \\
 & 30\% & 10.89 & 21.28 & 19.12 & 54.91 \\
 & 40\% & 17.06 & 37.84 & 32.33 & 49.70 \\
 & 50\% & 31.13 & 79.77 & 64.55 & 43.67 \\
\hline
\multirow{4}{*}{+MGAA}
 & 20\% & 7.42 & 11.93 & 11.22 & 60.66 \\
 & 30\% & 9.14 & 16.33 & 14.94 & 57.46 \\
 & 40\% & 13.09 & 27.68 & 23.80 & 52.93 \\
 & 50\% & 20.58 & 57.25 & 40.28 & 47.80 \\
\hline
\end{tabular}
\label{other_method}
\end{table}

\renewcommand{\arraystretch}{1.3}
\begin{table}[htbp]
\centering
\setlength\tabcolsep{8.5pt}
\caption{Performance comparison of different similarity measures.}
\begin{tabular}{c|ccc|c}
\hline
Distance & WikiText2 $\downarrow$ & PTB $\downarrow$ & C4 $\downarrow$ & Average $\uparrow$ \\
\hline
    Cosine & 24.54 & 66.86 & 50.22 & 44.59 \\
    Euclidean & 47.86 & 109.81 & 99.41 & 38.22 \\
    Manhattan & 41.89 & 98.26 & 75.09 & 39.36 \\
    N-Euclidean & 48.75 & 108.34 & 98.93 & 38.16 \\
\hline
\end{tabular}
\label{distance explor}
\end{table}
\renewcommand{\arraystretch}{1.0}

\indent \emph{4) Impact of the Calibration Dataset Type.}
We investigate the impact of various calibration datasets on the performance of the compressed LLaMA1 models at 50\% compression ratio. The final calibration data is selected from WikiText2, C4 \cite{c4}, and Alpaca datasets \cite{alpaca}, either individually or in combination. For combined datasets, we maintain a total sample size of 256 sequences (with equal distribution among datasets), each containing 512 tokens.

 We evaluate model performance using perplexity on WikiText2, PTB \cite{PTB}, and C4 datasets, as well as average accuracy across seven common reasoning datasets. The results are presented in Table \ref{Dataset type}. The experimental results illustrate that when using a single calibration dataset (e.g., WikiText2 or C4), the compressed model leads to optimal performance on the corresponding test set but underperforms on other tasks. For instance, using WikiText2 as the calibration data results in the lowest perplexity on the WikiText2 (17.09) but higher perplexities on PTB (91.39) and C4 (56.72). This domain-specific optimization occurs because the compression method adapts to the statistical properties of the calibration dataset. Compression method may overfit on the calibration data, resulting in poor generalization to other tasks. The utilization of mixed datasets significantly improves this issue to some extent. When maintaining a consistent data volume, employing a mixed dataset enables the compressed model to achieve more robust performance across multiple downstream tasks. The balance between language modeling and reasoning performance achieved by the mixed dataset underscores the importance of diverse and representative calibration data in model compression.

\indent \emph{5) Influence of the  Calibration Dataset Volume.}
We examine how calibration data volume affects model performance by varying both sequence length and sample quantity. Using mixed WikiText2 and Alpaca datasets at 50\% compression ratio, we conduct systematic experiments as shown in Table \ref{length and samples}. We observe that when the data volume is relatively small, increasing it leads to significant performance improvements. However, once the total data volume exceeds $512 \times 512$ (samples $\times$ length), the marginal benefits of further increases diminish. Furthermore, we observed that when the total data volume remains constant, such as in the cases of $256 \times 256$ versus $128 \times 512$, the compressed model exhibits comparable performance across multiple datasets. For instance, the perplexity on Wikitext2 is 24.71 and 24.53, while the average accuracy is 46.97 and 46.11, respectively. This suggests that the model is primarily determined by the total data size rather than the separate sequence length or sample count.

\indent \emph{6) Analysis of Hyperparameters.} 
In our  MGAA, we introduce a crucial hyperparameters $\alpha$. The parameter $\alpha$ controls the magnitude of compression ratio variations among different sublayers, where a larger $\alpha$ leads to greater differences in compression ratios between sublayers. 

\begin{table}[tb]
\centering
\setlength\tabcolsep{7pt}
\caption{Performance comparison across different calibration dataset types.}
\begin{threeparttable}
\begin{tabular}{c|ccc|c}
\hline
Dataset & WikiText2 $\downarrow$ & PTB $\downarrow$ & C4 $\downarrow$ & Average $\uparrow$ \\ 
\hline
WikiText2 & 17.09 & 91.39 & 56.72 & 43.08 \\
Alpaca & 67.21 & 77.71 & 45.00 & 49.19 \\
C4 & 65.24 & 79.41 & 28.16 & 48.43 \\
WikiText2-Alpaca & 22.02  & 58.57  & 43.33  & 47.55 \\
C4-Alpaca & 53.18 & 64.26 & 32.25 & 49.20 \\
C4-WikiText2 & 19.26 & 71.18 & 31.32 & 47.84 \\
ALL & 21.82 & 54.36 & 33.21 & 48.49 \\
\hline
\end{tabular}
\end{threeparttable}
\label{Dataset type}
\end{table}

\renewcommand{\arraystretch}{1.25}
\begin{table}[tb]
\centering
\setlength\tabcolsep{6pt}
\caption{Impact of sequence length and sample size on model performance.}
\begin{tabular}{c|c|ccc|c}
\hline
Samples & Length & WikiText2 $\downarrow$ & PTB $\downarrow$  & C4 $\downarrow$ & Average $\uparrow$ \\ 
\hline
\multirow{5}{*}{256} & 512 & 22.02 & 58.57 & 43.33 & 47.55 \\
\cline{2-6}
 & 256 & 24.71 & 71.88 & 45.03 & 46.97 \\
\cline{2-6}
 & 128 & 28.90 & 88.66 & 52.38 & 46.34 \\
\cline{2-6}
 & 64 & 36.78 & 118.87 & 63.97 & 43.95 \\
\cline{2-6}
 & 32 & 54.26 & 184.18 & 85.43 & 41.62 \\
\hline
2048 & \multirow{6}{*}{512} & 19.92 & 47.80 & 38.16 & 48.51 \\
\cline{1-1}\cline{3-6}
1024 & & 20.10 & 48.37 & 38.90 & 48.41 \\
\cline{1-1}\cline{3-6}
512 & & 20.84 & 51.58 & 40.37 & 47.99 \\
\cline{1-1}\cline{3-6}
256 & & 22.02 & 58.57 & 43.33 & 47.55 \\
\cline{1-1}\cline{3-6}
128 & & 24.53 & 64.66 & 48.41 & 46.11 \\
\cline{1-1}\cline{3-6}
64 & & 28.68 & 78.03 & 57.71 & 45.85 \\
\hline
\end{tabular}

\label{length and samples}
\end{table}

\renewcommand{\arraystretch}{1.25}
\begin{table}[htbp]
\centering
\setlength\tabcolsep{11.5pt}
\caption{The influence of hyperparameter $\alpha$ on the model performance.}
\begin{tabular}{c|ccc|c}
\hline
$\alpha$ & WikiText2 $\downarrow$ & PTB $\downarrow$ & C4 $\downarrow$ & Average $\uparrow$ \\
\hline
    0.10 & 29.63 & 74.49 & 60.45 & 43.49 \\
    0.20 & 26.58 & 69.34 & 54.49 & 44.22 \\
    0.25 & 25.54 & 66.91 & 52.06 & 44.56 \\
    0.30 & 24.88 & \textbf{65.14} & 50.73 & 44.37 \\
    0.35 & \textbf{24.54} & 66.86 & \textbf{50.22} & \textbf{44.59} \\
    0.40 & 24.58 & 67.90 & 50.23 & 44.54 \\
    0.45 & 25.09 & 71.44 & 51.28 & 44.55 \\
    0.50 & 25.70 & 76.22 & 51.69 & 44.14 \\
\hline
\end{tabular}
\label{alpha explor}
\end{table}

We perform a fine-grained exploration of $\alpha$ within the range [0.1, 0.5]. As shown in Table \ref{alpha explor}, the model's overall performance exhibits an initial improvement as $\alpha$ increases from 0.1 to 0.35. Specifically, when $\alpha = 0.35$, the model achieves optimal performance across multiple metrics, with the lowest perplexity on WikiText2 (24.54) and C4 (50.22) datasets, as well as the highest average accuracy (44.59) on downstream tasks. However, as $\alpha$ exceeds 0.35, we observe a consistent rise in perplexity scores across all datasets, accompanied by a decline in average accuracy. These findings suggest that small $\alpha$ values ($<$ 0.25) result in insufficient layer-wise compression ratio differentiation, which fails to fully leverage the inherent differences between layers. Conversely, while larger $\alpha$ values ($>$ 0.40) enhance the differentiation of compression ratios between layers, they may result in excessive compression in certain layers, thereby compromising the model's overall performance. Based on our comprehensive analysis of both perplexity metrics and downstream task performance, we recommend setting $\alpha = 0.35$ as the optimal configuration for model compression.

\begin{figure}[h]
  \centering
    \includegraphics[width=0.95\columnwidth]{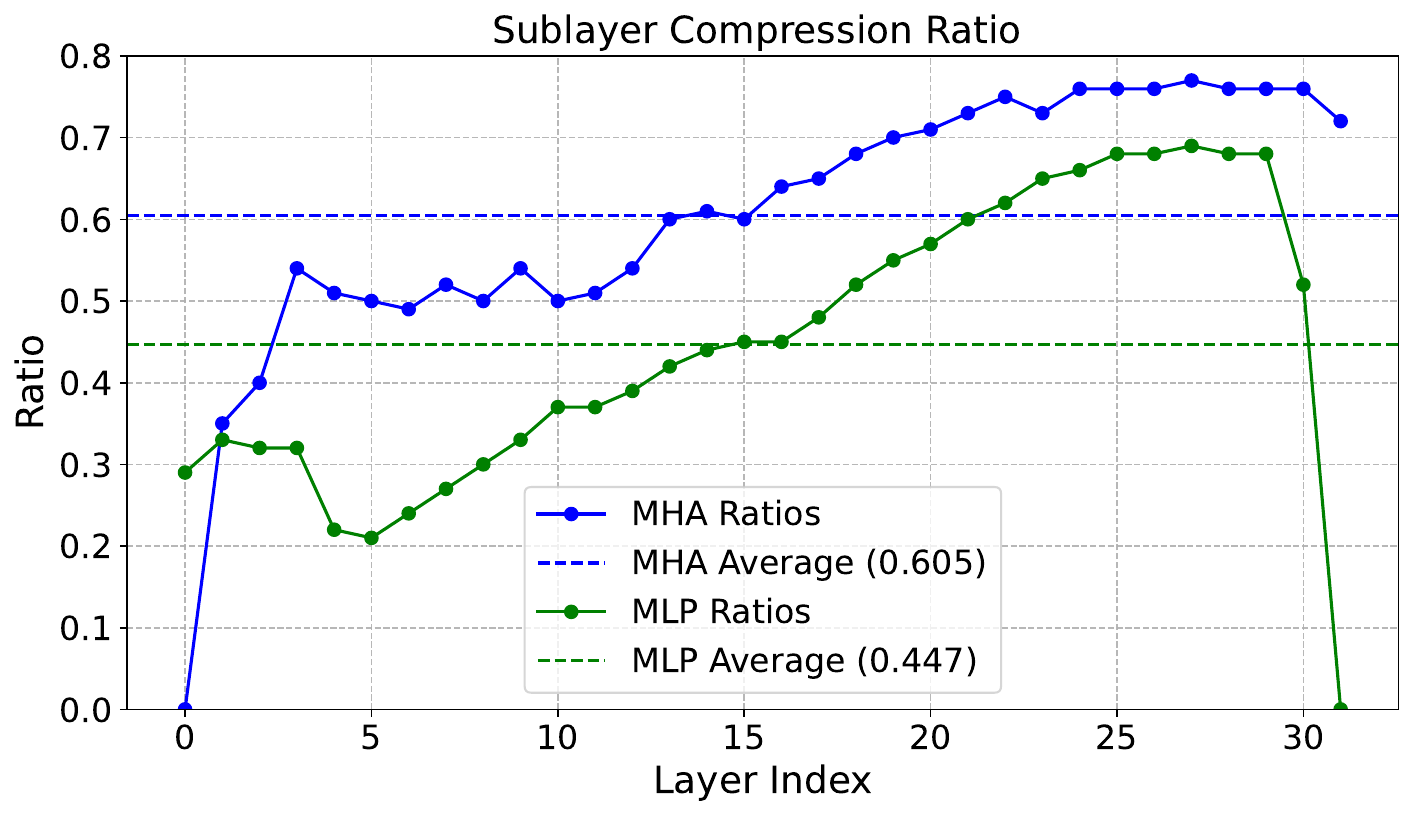}
  \caption{At an overall compression of 50\%, the sublayer compression of LLaMA1 model after the adjustment.}
  \label{ratio visualization1}
\end{figure}

\begin{figure}[h]
     \centering
    \includegraphics[width=0.95\columnwidth]{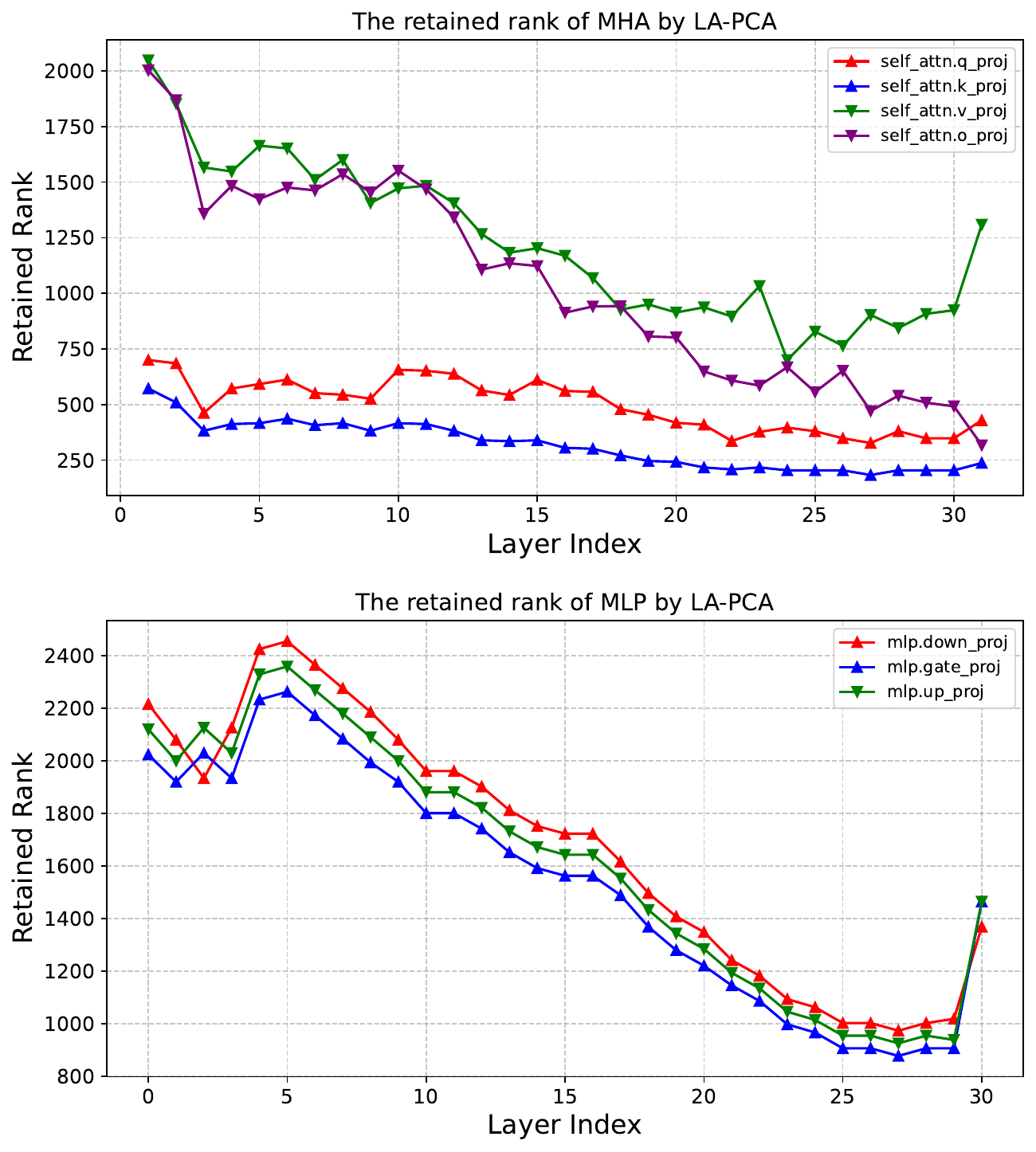}
  \caption{Under the 50\% compression ratio, the LLaMA1 model retains the ranks for its weight matrices.}
  \label{ratio visualization}
\end{figure}

\subsection{Visualization of the Compressed Model}
\indent \emph{1) Layer Compression Ratio.}
We have depicted the sublayer compression ratios of the LLaMA1 model at 50\% compression ratio in Figure \ref{ratio visualization1}. The visualization reveals distinctive compression patterns across different model depths and sublayers. 
Overall, both MLP and MHA sublayers demonstrate an increasing trend in compression ratios as network depth increases. This indicates that layer importance gradually decreases as the network deepens.
For MLP sublayers, we observe notably low compression ratios in the first several layers and the final layer, indicating their critical importance to model performance. The average compression ratio of MLP sublayers (0.447) is lower than the model's overall 50\% compression target, suggesting our method allocates more parameters to preserve MLP functionality. In contrast, MHA sublayers exhibit a higher average compression ratio of 0.605. This result indicates that MHA structure is more suitable for more aggressive compression, especially in the middle and later layers of the network.

\indent \emph{2) Retained Rank of Weight Matrix.}
We present the retained ranks of the weight matrices in the LLaMA1 model at a 50\% compression ratio in Figure \ref{ratio visualization}. Within the MHA sublayer, we observe distinct patterns in rank retention across different projection matrices: the $\mathbf{W}_{v}$ matrix consistently maintains the highest rank, while the $\mathbf{W}_{k}$ matrix retains the lowest. The query projection ($\mathbf{W}_{q}$) exhibits similar rank values to $\mathbf{W}_{k}$, both being substantially lower than the value and output projections
, which may be attributed to the computational characteristics of the attention mechanism. Notably, in the final MHA layer, both $\mathbf{W}_{v}$ and $\mathbf{W}_{o}$ matrices show significant changes in their retained ranks, highlighting the critical role of this layer in the model's overall functionality. For the MLP component, all three matrices maintain similar rank values throughout the network. This consistent pattern suggests these matrices contribute comparably to the model's functionality.

\begin{table*}[htbp]
\centering
\setlength\tabcolsep{3pt}
\caption{The performance improvement of AWSVD and PCA combined with our method MGAA. "SCQA" stands for the ScienceQA  dataset.} 
\begin{tabular}{c|c|ccccccccccc}
\hline
Ratio & Method & RealworldQA  & MMMU  & SCQA-img & SCQA-full & MME-C & MME-P & ChartQA & InfoVQA & AI2D & VizWiz & TextVQA  \\
\hline
 0\% & Dense & 57.91 &36.33 &70.25& 72.81 & 322.50 & 1522.72 & 54.80 &37.20 & 65.41 & 60.71 & 64.92 \\
\hline
\multirow{4}{*}{20\%}
 & AWSVD & 36.86 & 24.89 & 16.41 & 9.13 & 231.43 & 873.43 & 10.32 & 11.6 & 21.18 & 25.74 & 35.00 \\
 & PCA   & 47.63 & 27.33 & 48.59 & 50.65 & 259.29 & 1354.18 & 38.40 & 22.88 & 42.10 & 47.60 & 52.88\\
 & MGAA-AWSVD & 45.75 & 30.00 & 36.89 & 22.42 & 245.36 & 891.97 & 23.64 & 19.81 & 42.00 & 52.12 & 42.74 \\
 & MGAA-PCA  & \textbf{48.76} & \textbf{33.33} & \textbf{57.41} & \textbf{61.75} & \textbf{265.71} & \textbf{1445.39} & \textbf{44.92} & \textbf{27.45} & \textbf{54.73} & \textbf{61.18} & \textbf{55.27}\\
\hline
\multirow{4}{*}{30\%}
 & AWSVD & 6.27 & 26.22 & 0.00 & 0.05 & 17.50 & 83.68 & 1.16 & 2.03 & 1.20 & 16.01 & 7.02 \\
 & PCA   & 36.47 & 26.78 & 28.81 & 36.36 & 237.14 & 1260.86 & 26.48 & 17.89 & 21.66 & 9.34 & 45.06\\
 & MGAA-AWSVD & 29.67 & 26.22 & 13.68 & 7.33 & 190.36 & 559.71 & 4.24 & 3.39 & 12.37 & 37.71 & 13.71 \\
 & MGAA-PCA   & \textbf{38.95} & \textbf{28.56} & \textbf{33.76} & \textbf{42.84} & \textbf{251.07} & \textbf{1327.67} & \textbf{35.48} & \textbf{22.38} & \textbf{31.54} &\textbf{ 56.53} & \textbf{47.72}\\
\hline
\end{tabular}
\label{llava}
\end{table*}

\subsection{Experiments on Multi-Modal Model}
In this subsection, we further verify the effectiveness of our MGAA on multi-model LLMs.

\indent \emph{1) The Backbones.} We evaluate our approach on LLaVA-1.6 \cite{llava}, a state-of-the-art multi-modal model built upon Vicuna 7B \cite{Vicuna}. The model architecture consists of a CLIP-ViT-L-336px visual encoder with an MLP projection layer and a Vicuna language model. Trained on 1.2 million diverse image-text pairs, LLaVA-1.6 achieves superior performance across multiple benchmarks.

\indent \emph{2) Calibration Data.}
In our experimental setup, we employ the VQAv2 dataset \cite{vqa2v} as our calibration dataset. VQAv2 \cite{vqa2v} stands as one of the most seminal datasets in multimodal research, comprising approximately 1.2 million question-answer pairs. The dataset encompasses a diverse spectrum of tasks, ranging from fundamental object recognition and enumeration to sophisticated queries that demand complex reasoning capabilities and comprehensive commonsense knowledge, including but not limited to causal inference and logical deduction. This rich diversity in task complexity and reasoning requirements makes VQAv2 an ideal candidate for evaluating and calibrating multimodal comprehension capabilities. We extracted 256 sequences with a padding length of 3000 from the VQAv2 dataset as calibration samples.

\indent \emph{3) Evaluation.}
To comprehensively evaluate the model's performance across diverse scenarios, we conduct experiments on nine representative multimodal benchmarks: 
1) \textbf{MMMU} \cite{mmmu}: A comprehensive multi-modal benchmark containing tasks across professional domains including medicine, art, and engineering, designed to evaluate specialized knowledge and reasoning.
2) \textbf{ScienceQA} \cite{scienceqa}: An educational dataset focusing on K-12 science curriculum, testing scientific understanding and reasoning abilities at various grade levels.
3) \textbf{MME} \cite{mme}: A multi-modal evaluation benchmark designed to assess models' capabilities in perception and cognitive reasoning across diverse scenarios.
4) \textbf{ChartQA} \cite{chartqa}: A specialized dataset focusing on chart comprehension and interpretation, testing the ability to extract and reason about graphical information.
5) \textbf{InfoVQA} \cite{infovqa}: A collection of information-seeking visual questions designed to evaluate models' ability to provide accurate and relevant responses to practical queries.
6) \textbf{AI2D} \cite{ai2d}: A diagram-focused dataset testing scientific reasoning through visual interpretation of technical and educational diagrams.
7) \textbf{VizWiz} \cite{vizwiz}: A unique dataset comprising real-world visual questions collected from visually impaired individuals, representing authentic accessibility challenges.
8) \textbf{TextVQA} \cite{textqa}: A dataset focusing on text reading and comprehension within natural images, testing both visual and textual understanding.
9) \textbf{RealworldQA}: A practical application-oriented dataset designed to evaluate models' performance in real-world scenarios and everyday problem-solving situations. This diverse suite of benchmarks enables comprehensive assessment of the model's capabilities across various domains, task types, and complexity levels. All evaluations are conducted using lmms-eval \cite{lmms_eval} with a batch size of 1 on a single A100 GPU (80GB).

\indent \emph{4) Implementation Details.}
Although LLaVA is fine-tuned from Vicuna, we observe distinct compression characteristics in our experiments. The LLaVA model shows heightened sensitivity to compression in the earlier layers responsible for visual-linguistic integration. Based on this observation, we preserved the first four layers without compression and redistributed the compression load to the remaining layers while maintaining the target overall compression ratio. This strategy effectively preserved the model's multimodal processing capabilities.

\indent \emph{5) Results Analysis.}
In Table \ref{llava}, we analyze the experimental results of two representative methodologies: PCA for feature decomposition and AWSVD for weight decomposition techniques. Our results demonstrate that the MGAA method consistently improves performance across multiple evaluation metrics and datasets. At 20\% compression ratio, MGAA-PCA achieves significant improvements over baseline PCA, with scores increasing from 27.33 to 33.33 on MMMU and from 48.59 to 57.41 on SCQA-img. When MGAA was integrated with AWSVD (MGAA-AWSVD), the method shows particular strength in handling challenging scenarios, notably achieving a score of 29.67 on RealworldQA at 30\% compression ratio while maintaining stable performance across other metrics. The effectiveness of our approach is further validated by consistent improvements across ChartQA, InfoVQA, and AI2D benchmarks. These comprehensive results demonstrate that the MGAA method effectively enhances the model performance.

The multi-modal model exhibits notably higher compression sensitivity compared to traditional language models. Even at relatively low compression ratios, we observe significant performance degradation. For instance, at 30\% compression ratio, both baseline PCA and our proposed MGAA method experience considerable performance drops, though MGAA demonstrates better performance retention. This increased sensitivity to compression suggests a lower degree of redundancy in the multi-modal architecture, indicating that parameters previously redundant in the language model have been effectively utilized through fine-tuning on multi-modal tasks, thereby acquiring visual processing capabilities. 

The above observations provide valuable insights into the model's parameter efficiency and the transformation of network capacity during multimodal adaptation. The reduced compressibility of the model after multimodal fine-tuning suggests that the originally redundant parameters have been repurposed to support visual understanding tasks, demonstrating the model's efficient utilization of its parametric capacity in handling multimodal information processing.

\section{Conclusion}
In this work, we introduce Multi-Granularity Adaptive Allocation (MGAA), a novel parameter allocation method specifically designed for low-rank approximation of LLMs. Our approach implements an adaptive global parameter allocation strategy while maintaining the overall target compression ratio. We utilize sublayer-level importance metric based on input-output similarity to allocate compression ratios across different sublayers. Additionally, we also propose an energy-balanced adaptive allocation method, which allocate the retained parameter for weight matrices within the same sublayer. Empirical results demonstrate that MGAA can be combined with various low-rank approximation methods to significantly enhance the performance of compressed model across multiple compression ratios. This fully demonstrates the superior performance and broad applicability of the MGAA, and also paves the way for its potential to integrate with more advanced low-rank approximation methods. We believe our work contributes to a better understanding of adaptive compression strategies for LLMs and  provides valuable insights for research in efficient, high-performance model compression techniques.

\ifCLASSOPTIONcaptionsoff
  \newpage
\fi




\bibliography{reference}

\vfill


\end{document}